\documentclass{article}
\usepackage[preprint]{neurips_2026}
\usepackage[utf8]{inputenc}
\usepackage[T1]{fontenc}
\usepackage{hyperref}
\usepackage{url}
\usepackage{url}
\usepackage{booktabs}
\usepackage{amsfonts}
\usepackage{nicefrac}
\usepackage{microtype}
\usepackage{xcolor}
\usepackage{listings}
\usepackage{enumitem}
\usepackage{subcaption}
\usepackage{wrapfig}

\usepackage{tcolorbox}

\usepackage{tikz}
\usetikzlibrary{arrows.meta, positioning, fit, calc, backgrounds, shapes.geometric}
\usepackage{multirow} 
\usepackage{amsmath}
\usepackage{todonotes}
\usepackage{ragged2e} 
\usepackage{caption}
\usepackage{booktabs}
\usepackage{enumitem} % Added for better list control
\title{ COMPOSE: Composing Future Theorems from Citations and Formal Structure
}

\author{%
  David Busbib \qquad \quad
  Michael Werman \\ \\
  Hebrew University of Jerusalem \\
}
\newif\ifdraft
\drafttrue
\ifdraft
\newcommand{\david}[1]{{\color{red}[\textbf{David:} #1]}}

\newcommand{\mike}[1]{{\color{cyan}[\textbf{Mike:} #1]}}

\else
\newcommand{\david}[1]{}
\newcommand{\mike}[1]{}

\fi

\begin{document}

\maketitle

% \begin{figure}[t]
% \centering
% \input{figures/training}
% \end{figure}

\begin{abstract}
A plausible future mathematical claim must satisfy two constraints: it should follow the direction of prior work and respect the formal dependencies that constrain what can validly follow. Existing approaches typically model only one of these sources, producing claims that are either weakly grounded or insufficiently motivated. We introduce grounded future mathematical generation, where the goal is to generate a plausible future theorem-like claim for an anchor paper using two complementary sources of context: its scientific citation graph and aligned formal theorem dependency graph. To address this setting, we propose COMPOSE, a dual-graph framework that conditions a language model on both scientific citation context and formal theorem structure. To support this setting, we construct a dataset of 108K paired scientific-formal graph examples from arXiv and Mathlib, together with a benchmark of 47K future papers from 2024--2025. Experiments show that COMPOSE outperforms strong baselines on retrieval to real future papers and achieves the best overall performance under LLM-judge evaluation, producing more grounded and mathematically richer outputs. These results show that future mathematical generation benefits from combining scientific context with formal structure. Project page is available at \url{https://david-busbib.github.io/COMPOSE-page/}.
\end{abstract}

\section{Introduction}
\label{sec:intro}

% \begin{figure}[t]
% \centering
% \input{figures/arxiv_over}
% \caption{...}
% \end{figure}

Large language models have shown growing promise in anticipating future scientific contributions from existing knowledge across many domains~\citep{si2024llmsgeneratenovelresearch,lu2024aiscientistfullyautomated,wang2026learningpredictfuturealignedresearch}. However, in the mathematical domain, a plausible future result must both extend the research trajectory of prior work~\citep{lakatos1976proofs,davies2021advancing} 
but also respect the logical dependencies that determine what can validly follow from existing definitions, lemmas, and theorems~\citep{yang2023leandojo,uskuplu2025knowtexvisualizingmathematicaldependencies,cabral2025proofflowdependencygraphapproach}. Thus, the task is not only to produce fluent mathematical text, but to generate a result that is both scientifically meaningful and logically grounded.

This challenge is reflected in two complementary sources of mathematical knowledge: scientific papers, which capture how ideas develop across the literature and indicate which directions are worth pursuing~\citep{azher2025futuregenragbasedapproachgenerate,baek2025researchagentiterativeresearchidea,heyueya2026giantsgenerativeinsightanticipation}, and formal theorem libraries such as Lean’s Mathlib, which encode mathematical results together with explicit dependency structure.
Combining the two is non-trivial because the same content appears differently in each source, and because the two sources are organized by fundamentally different relational structures~\citep{gao2024semantic,zhang2025autoformalizationwildassessingllms,wu2022autoformalizationlargelanguagemodels}. Yet existing approaches tackle only one side: models trained on scientific text alone can identify plausible research directions but may generate claims that are not grounded in formal mathematical dependencies~\citep{Wang_2024,radensky2026humanllmcompoundscientificideation, baek2025researchagentiterativeresearchidea,heyueya2026giantsgenerativeinsightanticipation};
theorem-proving systems~\citep{wang2025kiminaproverpreviewlargeformal,first2023baldurwholeproofgenerationrepair,yang2023leandojo} reason over formal structure but lack the surrounding scientific context needed to anticipate which directions are most promising; 
and structure-based methods typically model only one of these sources, using either citation structure~\citep{an2021enhancingscientificpaperssummarization} or theorem dependencies~\citep{wang2025ariaagentretrievaliterative,zhu2026leanarchitectautomatingblueprintgeneration}. As a result, existing work still lacks a method that combines both sources for grounded future mathematical generation.

\begin{figure}[t]
\centering
\input{figures/training}
\label{fig:pipeline}
\vspace{-1.5em}
\end{figure}

In this work, we propose \textit{COMPOSE}, a system for generating plausible future mathematical claims from an anchor paper. This is carried out by jointly modeling scientific context and formal mathematical structure. To represent these two sources,  we construct two complementary graphs: \textit{(i)} a scientific graph built from the anchor paper, selected cited papers, and extracted theorems, and
 \textit{(ii)} a formal graph built by aligning those theorems to corresponding formal theorems and their dependency structure. These two graphs are encoded separately and fused into a shared representation that conditions an LLM decoder on both the scientific direction of the papers and the formal structure of the underlying mathematics.
Figure~\ref{fig:pipeline} provides an overview of the full pipeline.
To learn this representation effectively, training proceeds in two stages: we first optimize the graph encoders and fusion module with alignment objectives, and then fine-tune the decoder for graph-conditioned generation. The model then generates a theorem-like mathematical text that represents a plausible contribution in the same research direction, grounded in both the evolution of ideas across papers and the dependency structure of formal mathematics.

To evaluate this setting, we construct a benchmark of 108K paired samples from arXiv mathematics using training papers collected from the pre-2023 corpus, together with a test set of 47K papers from late 2024 and 2025.

Empirically, \textsc{COMPOSE} achieves the strongest retrieval performance in our main evaluation setting and performs best overall under LLM-judge evaluation, with ablations confirming that both graph sources contribute.

We make the following contributions:
\begin{itemize}

\item We study grounded future mathematical generation, where the goal is to predict plausible future mathematical claims using both scientific citation context and formal theorem dependencies.
\item We construct a dataset of 108K paired scientific-formal graph samples from arXiv and Mathlib, using an informal-to-informal alignment strategy via FrenzyMath, and establish a benchmark of 47K future papers from 2024-2025 for evaluation.
\item We propose \textsc{COMPOSE}, a system for grounded future mathematical generation that combines scientific citation context with formal theorem structure to generate plausible future mathematical claims.

\end{itemize}

\section{Related Work}

\paragraph{Future Scientific Generation.}
Prior work studies future scientific generation in several ways.
One line of work generates future research directions from prior literature using retrieval or iterative retrieval-and-reasoning over related papers~\citep{azher2025futuregenragbasedapproachgenerate,baek2025researchagentiterativeresearchidea,radensky2026humanllmcompoundscientificideation}. Another line of work uses citation structure or temporal supervision for the task~\citep{wang2026learningpredictfuturealignedresearch,Wang_2024}. GIANTS~\citep{heyueya2026giantsgenerativeinsightanticipation} uses citation-based scientific context to guide generation, although it targets future scientific summaries rather than theorem-like claims. Some work also explores structure-aware scientific generation using external graphs or chain-of-ideas frameworks~\citep{gao2025goaienhancingaistudents,xiong2024improvingscientifichypothesisgeneration}. Across these settings, existing systems rely on textual, citation-based, or structured scientific context, without combining the formal theorem structure that constrains what can validly follow from prior results.

\paragraph{Formal Mathematical Reasoning.} 

Formal proof libraries such as Lean's Mathlib~\citep{mathlib2019}, Isabelle's Archive of Formal Proofs~\citep{paulson1994isabelle}, and Coq-based libraries~\citep{barras1999coq,rudnicki1992overview} encode large collections of verified theorems together with their formal dependency structure~\citep{li2024surveydeeplearningtheorem}. This dependency structure is not just a record of what has been proved; it also helps determine what can be proved next. Retrieval-augmented and premise-selection systems use this structure to identify relevant premises for proof search, as in ReProver~\citep{yang2023leandojo} and related work~\citep{alemi2017deepmathdeepsequence,wang2017premiseselectiontheoremproving,mikuła2024magnushammertransformerbasedapproachpremise}. Formal theorem provers also use this library context to guide proof generation and search, including systems such as Baldur~\citep{first2023baldurwholeproofgenerationrepair} and DeepSeek-Prover~\citep{xin2024deepseekproveradvancingtheoremproving}, as well as more recent work~\citep{wang2025kiminaproverpreviewlargeformal,dong2025stpselfplayllmtheorem}. Beyond proof search, formal libraries have also been used to conjecture what should be proved in the first place: Lemmanaid~\citep{alhessi2026lemmanaidneurosymboliclemmaconjecturing} proposes new lemmas by drawing analogies across existing theories, while related work studies conjecture generation, verification, and blueprint construction from library context~\citep{onda2025leanconjecturerautomaticgenerationmathematical,kasaura2025discoveringnewtheoremsllms,liu2026mathliblemmafolklorelemmageneration,zhu2026leanarchitectautomatingblueprintgeneration}. Together, these systems show that formal dependency structure can inform not only how to prove but also what to prove. However, they operate entirely within the formal world, requiring formal input and producing formal output, without modeling the informal mathematical context found in research papers.

\paragraph{Scientific Informal-Formal Alignment.}
In our setting, the alignment problem is to match an informal theorem from a research paper to its corresponding theorem in Mathlib. A natural approach is to translate informal mathematics written in natural language into formal proof-assistant syntax. Early work used large language models to map informal theorems into formal representations~\cite{wu2022autoformalizationlargelanguagemodels,azerbayev2023proofnetautoformalizingformallyproving,lu2024processdrivenautoformalizationlean4}. Later work improved Autoformalization with stronger training, retrieval-based methods, and math-specialized models, including Kimina~\citep{wang2025kiminaproverpreviewlargeformal}, Draft, Sketch, and Prove~\citep{jiang2023draftsketchproveguiding}, and ProofBridge~\citep{jana2026proofbridge}. Despite these advances, coverage on real paper-level mathematics remains limited because of the large vocabulary gap between informal exposition and formal syntax~\cite{zhang2025autoformalizationwildassessingllms}. As an alternative, FrenzyMath~\citep{gao2024semantic_mathlib4} takes a different route by providing human-written informal descriptions for over 136K Mathlib theorems, enabling informal-to-informal retrieval as a pragmatic alignment strategy that avoids formalization altogether.
Recent work also suggests that linking informal theorem to formal libraries benefits from stronger retrieval and intermediate structure~\citep{lu2024processdrivenautoformalizationlean4,jiang2023draftsketchproveguiding,jana2026proofbridge}.
Unlike these works, we do not treat informal-formal alignment as the primary objective. Instead, we use the resulting aligned structure to support grounded generation of future mathematical claims.

\section{Method}
Given an anchor paper, the goal is to generate a plausible future mathematical claim using its scientific and matched formal context. Our approach has three components. First, we construct a dataset that links scientific papers with formal theorem graphs~\ref{dataset}. Second, we introduce \textsc{COMPOSE}, a dual-encoder architecture that jointly models scientific and formal mathematical structure~\ref{eq:model_name}. Third, we present our two-stage training~\ref{two_stage_trian}.

\paragraph{Inference overview.}
At inference time, given an anchor paper, the model constructs two graphs: a scientific citation graph $G_s$, built from the paper and its contextually relevant cited papers, and a formal theorem dependency graph $G_f$, built from Mathlib theorems aligned to extracted mathematical theorems. Each graph is independently encoded by a dedicated GNN encoder. The two representations are then fused via cross-attention. The fused graph embeddings are passed as conditioning to our language model decoder, which generates a free-text theorem proposal.

\subsection{Data Curation and Construction}\label{dataset}

Training our dual-graph model requires paired examples linking scientific context with formal theorem-dependency structure. We construct this dataset by linking mathematical papers collected via S2ORC~\citep{lo2020s2orcsemanticscholaropen} with formal theorems from Mathlib~\citep{mathlib2019}.

On the scientific side, we begin with an anchor paper and construct outgoing citation edges for up to two hops, retaining at most 5 cited papers at the first hop and at most 3 cited papers per node for the second hop. Cited papers are selected by citation-context relevance, favoring citations that are discussed in important parts of the paper or used directly in its proofs or main results, rather than by including the full citation list; full selection details are  in Appendix~\ref{app:citation_subgraph}. The graph contains {\it abstract} nodes for all included papers, together with {\it theorem} nodes extracted from papers in the first two citation hops. Each {\it theorem} node is connected either to its corresponding {\it abstract} node or to a parent {\it theorem} node. We use typed edges to distinguish citation links, paper-to-theorem links, and theorem-to-theorem links, allowing the graph to capture both external and internal citation context. For each example, we also identify a target paper: a paper that cites the anchor and was published after it, excluded from the input graph. Its main mathematical claim serves as the generation target.

% Move this block to the text that appears in the right column
\begin{wrapfigure}[13]{r}{0.3\textwidth}
\vspace{-10.0pt}
  \centering
  % Alignment pipeline: vertical, compact
\begin{tikzpicture}[
  font=\scriptsize,
  box/.style={rounded corners=3pt, draw, inner sep=4pt, text width=3.0cm, align=center},
  arr/.style={-{Stealth[length=3pt]}, line width=0.8pt, gray!60},
  lbl/.style={font=\tiny\itshape, text=gray!50!black, align=center},
]

\node[box, fill=orange!8, draw=orange!60!black] (informal) at (0,0) {%
  \textbf{Informal}\\[1pt]
  {\tiny ``...primes dividing $|G|$ is infinite''}
};

\node[box, fill=gray!6, draw=gray!50, below=0.5cm of informal] (frenzy) {%
  \textbf{FrenzyMath Match}\\[1pt]
  {\tiny ``Infinitude of Primes''}\\[1pt]
  {\tiny \texttt{Nat.infinite\_setOf\_prime}}
};

\node[box, fill=purple!6, draw=purple!50!black, below=0.5cm of frenzy] (mathlib) {%
  \textbf{Mathlib Root}\\[1pt]
  {\tiny \texttt{:\{p | Prime p\}.Infinite}}
};

\draw[arr] (informal.south) -- (frenzy.north)
  node[midway, right=2pt, lbl] {e5-mistral-7b};
\draw[arr] (frenzy.south) -- (mathlib.north)
  node[midway, right=2pt, lbl] {+deps};

\end{tikzpicture}
  \caption{Informal-to-informal alignment.}
  \label{fig:alignment}
\end{wrapfigure}
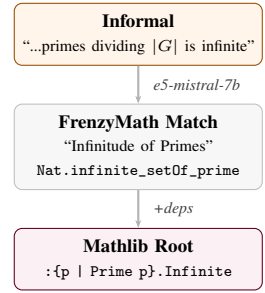

On the formal side, we align extracted theorems to Mathlib theorems using dense retrieval over the FrenzyMath corpus~\citep{gao2024semantic_mathlib4}, which provides informal descriptions for 140k Mathlib theorems (see Figure~\ref{fig:alignment}).
For each extracted theorem, we keep the top retrieved theorem only if its similarity is above a threshold $\tau$; weak matches are discarded. Each retained theorem serves as the root of the local formal dependency graph, which is expanded for a fixed number of hops using directed theorem-dependency edges extracted from Mathlib with LeanDojo~\citep{yang2023leandojo}. Root nodes are marked with a distinct node type, while the remaining nodes represent dependency neighbors. A comparison with alternative alignment strategies and additional details about $\tau$ is provided in Appendix~\ref{app:alignment}.

Starting from a corpus of 500K mathematical papers published between 2000 and 2023, this process yields approximately 108K paired paper-theorem subgraphs.
In addition, we define a temporally held-out benchmark of 47K mathematical papers from late 2024 and 2025, using their {\it abstract} and extracted main claim, as future target texts for evaluation. 
Full details of the alignment procedure, target construction, filtering steps, and dataset statistics are provided in Appendix~\ref{app:dataset}.

\subsection{COMPOSE} \label{eq:model_name}

We first introduce notation, then present the encoder, fusion, and decoder components.

Formally, each example is represented as
\[
x = (G_s, G_f, \mathcal{P}), \qquad
G_s = (V_s,E_s), \qquad
G_f = (V_f,E_f),
\]
where \(x\) is the input context and \(y=(y_1,\dots,y_T)\) is the target output sequence.

The scientific graph contains two types of nodes:
\[
V_s=\{v^{\mathrm{abs}}_j\}_{j=1}^n \cup \{u^{\mathrm{thm}}_i\}_{i=1}^m
,
\]
 where \(v^{\mathrm{abs}}_j\) denotes an abstract node corresponding to a paper in the scientific graph, and \(u^{\mathrm{thm}}_i\) are extracted theorem nodes. The edge set \(E_s\) contains directed citation edges between abstract nodes and structural edges connecting each theorem node either to its corresponding abstract node or to a parent theorem node.

The formal graph contains theorem nodes,
\(
V_f=\{v^{\mathrm{thm}}_\ell\}_{\ell=1}^k,
\)
where each node corresponds to a Mathlib theorem, and edge set \(E_f\) is given by dependency relations extracted from Mathlib.

Let \(V_s^{\mathrm{thm}}=\{u^{\mathrm{thm}}_i\}_{i=1}^m\) denote the set of theorem nodes in $G_s$.
The alignment set is defined as
\[
\mathcal{P}\subseteq V_s^{\mathrm{thm}}\times V_f,
\]
where each pair \((u^{\mathrm{thm}}, v^{\mathrm{thm}})\in\mathcal{P}\) links an informal theorem node in \(G_s\) to its matched formal theorem node in \(G_f\).

Each graph is processed by a dedicated GNN encoder.  The two encoders share the same overall message-passing architecture~\cite{gilmer2017neuralmessagepassingquantum} but differ in their node type and input initialization. In $G_s$, nodes are heterogeneous, with abstract and theorem nodes initialized from E5 embeddings ~\citep{wang2024textembeddingsweaklysupervisedcontrastive}. In $G_f$, nodes correspond to formal Lean theorems and are initialized from DeepSeek-Math embeddings~\citep{shao2024deepseekmathpushinglimitsmathematical} of theorem signatures, chosen for their stronger representation of formal mathematical content. Both encoders use directed, edge-type-specific message passing, with incoming and outgoing messages aggregated separately. The resulting updates are combined with a learned gated residual connection to mitigate over-smoothing, followed by layer normalization. For each node, the GNN produces a structural vector that is concatenated with the node's original text embedding, yielding contextualized representations $\mathbf{h}_i^s$ and $\mathbf{h}_j^f$ that capture both semantic content and graph structure. Full architectural details are  in Appendix~\ref{app:gnn}.

Since $G_s$ and $G_f$ operate in different representation spaces, we project $\mathbf{h}_i^s$ and $\mathbf{h}_j^f$ into a shared latent space using the same learned MLP. Type embeddings are then added after projection to distinguish nodes from $G_s$ and $G_f$. We apply a single bidirectional cross-attention block between the two node sets, with separate Q/K/V projections for the scientific-to-formal and formal-to-scientific directions, and residual connections with layer normalization applied after each directional update. The resulting fused node representations are denoted by \(\{\tilde{\mathbf{h}}_i\}_{i=1}^{|V_s|+|V_f|}\).

The fused node representations \(\{\tilde{\mathbf{h}}_i\}\) are provided as graph inputs to a pretrained math-specialized language model decoder adapted with LoRA~\cite{hu2021loralowrankadaptationlarge}. Cross-attention layers are inserted at regular intervals, covering approximately 20\% of the decoder layers. In these layers, the decoder hidden states serve as queries, while the fused graph node representations, projected to the decoder hidden dimension, serve as keys and values. This allows each generation step to attend over the full set of fused graph nodes, conditioning the output sequence \(y\) on both the scientific context of \(G_s\) and the formal structure of \(G_f\).

\subsection{Two-Stage Training }\label{two_stage_trian}

We train the model in two stages. The first learns graph representations that capture both within-graph structure and alignment between the scientific and formal domains, without using the decoder. The second adds the decoder and fine-tunes the full model for generation while preserving graph conditioning.

In the first stage, we train the scientific and formal graph encoders, together with the bridge MLP and fusion module, while keeping the pretrained text embeddings frozen. The resulting graph modules are then used to initialize the second stage. The first-stage objective is:

\begin{equation}
    \mathcal{L}_{\mathrm{stage}_1} = \lambda_{link}\mathcal{L}_{link} + \lambda_{align} \mathcal{L}_{align} + \lambda_{cross} \mathcal{L}_{cross}
\end{equation}

To ensure both encoders capture graph structure rather than collapsing to uniform representations, we supervise them with a link prediction loss. For each graph \(G \in \{G_s, G_f\}\), we define \(a_{ij}=1\) if \((v_i,v_j)\in E\) and \(a_{ij}=0\) for non-edges , and optimize
\begin{equation}
\mathcal{L}_{link} =
-\mathbb{E}\left[
a_{ij}\log\sigma(\mathbf{h}_i \cdot \mathbf{h}_j)
+
(1-a_{ij})\log\left(1-\sigma(\mathbf{h}_i \cdot \mathbf{h}_j)\right)
\right].
\end{equation}

where \(\mathbf{h}_i\) and \(\mathbf{h}_j\) denote representations in the graph under consideration, and $\sigma$ is the sigmoid function.

To encourage the fused graph representation to align with the scientific and formal content of the target, we use a contrastive alignment loss~\citep{oord2019representationlearningcontrastivepredictive}. Let
\begin{equation}
\tilde{\mathbf{h}}_G = \mathrm{AttnPool}(\{\tilde{\mathbf{h}}_i\}_{i=1}^{|V_s|+|V_f|})
\end{equation}
denote the fused graph representation obtained by attention pooling over the fused node representations.

We construct a positive target representation \(\mathbf{e}^{+}\) from the held-out target paper by combining the embeddings of its abstract and main mathematical claim, after projecting both into the shared latent space. For each example, we construct negative target representations \(\mathcal{B}^{-}\) from papers that don't cite  papers in the current scientific subgraph, together with their main mathematical claim.

This objective encourages the fused graph representation to be close to the true held-out target and far from unrelated target papers. We then optimize
\begin{equation}
\mathcal{L}_{align}
=
-\log
\frac{
\exp(\mathrm{sim}(\tilde{\mathbf{h}}_G,\mathbf{e}^{+})/\tau)
}{
\exp(\mathrm{sim}(\tilde{\mathbf{h}}_G,\mathbf{e}^{+})/\tau)
+
\sum_{\mathbf{e}^{-}\in\mathcal{B}^{-}}
\exp(\mathrm{sim}(\tilde{\mathbf{h}}_G,\mathbf{e}^{-})/\tau)
}
\end{equation}
where \(\mathrm{sim}(\cdot,\cdot)\) denotes cosine similarity.

Finally, to bridge the informal and formal spaces, $\mathcal{L}_{cross}$ aligns each informal theorem node $u_i^{\mathrm{thm}}\in V_s$ with its matched theorem node $v_j^{\mathrm{thm}} \in V_f$, where $(u_i^{\mathrm{thm}}, v_j^{\mathrm{thm}})\in\mathcal{P}$. For each positive pair, we sample a set of random negative theorem nodes $\mathcal{N}_i \subset V_f$. We then optimize
\begin{equation}
    \mathcal{L}_{cross}
    =
    -\frac{1}{|\mathcal{P}|}
    \sum_{(u_i^{\mathrm{thm}}, v_j^{\mathrm{thm}})\in\mathcal{P}}
    \log
    \frac{\exp(\hat{\mathbf{h}}_i^s \cdot \hat{\mathbf{h}}_j^f / \tau)}
    {\exp(\hat{\mathbf{h}}_i^s \cdot \hat{\mathbf{h}}_j^f / \tau) + \sum_{v_l^{\mathrm{thm}}\in\mathcal{N}_i}\exp(\hat{\mathbf{h}}_i^s \cdot \hat{\mathbf{h}}_l^f / \tau)}.
\end{equation}

where $\tau$ is the temperature hyperparameter.

In the second stage, we add a pretrained math-specialized decoder adapted via LoRA~\citep{hu2021loralowrankadaptationlarge} and fine-tune the full model for graph-conditioned generation using two objectives: an autoregressive generation loss and a graph margin loss that prevents the decoder from ignoring the graph context. The autoregressive loss is:
\begin{equation}
\mathcal{L}_{CE} = -\sum_{t=1}^{T} \log p(y_t \mid y_{<t}, G_s, G_f).
\end{equation}

To prevent the decoder from ignoring the graph context and relying purely on language patterns, we additionally use a graph margin loss \(\mathcal{L}_{margin}\) that requires the model to assign lower generation loss to the correct graph pair \((G_s, G_f)\) than to a mismatched pair \((G_s', G_f')\) sampled from a different citation graph within the same mathematical domain.

\begin{equation}
    \mathcal{L}_{margin} = \max\left(0,\, \mathcal{L}_{CE}(G_s, G_f) - \mathcal{L}_{CE}(G_s', G_f') + \delta\right)
\end{equation}

The  second stage objective is   

\begin{equation}
    \mathcal{L}_{\mathrm{stage}_2} = \lambda_{ce}\, \mathcal{L}_{CE} + \lambda_{margin}\, \mathcal{L}_{margin}
\end{equation}
Additional alignment and training details are provided in Appendix~\ref{app:gnn}.

\section{Experiments}
% \vspace{-0.8em}
\subsection{Experimental Setup}
% \vspace{-0.8em}
\textbf{Data and Temporal Split.}
The 108K paired scientific-formal graph samples are split 80/10/10 into training, validation, and test sets. To evaluate temporal generalization, we construct a future-paper benchmark from a corpus of 47K mathematics papers published in late 2024-2025. We retain only test-set anchor papers that are cited by at least one paper in the 47K future-paper pool, yielding 2K test samples. For each anchor, future papers that cite it are treated as targets.

\textbf{Retrieval Evaluation.}
Existing mathematical embedding models cluster future papers so tightly that the similarity gap between true target papers and randomly sampled unrelated papers is too small to reliably distinguish generation methods (see Appendix~\ref{app:embedding_analysis}). We therefore fine-tune a DeepSeek-Math embedding model on the 47K benchmark papers using a margin-ranking objective that pulls together papers with overlapping citation context and pushes apart unrelated papers. The resulting fixed model is then used as a shared retrieval evaluator for all compared methods.

\textbf{Metrics.}
We report Tgt-Sim, the mean cosine similarity between the generated text and its corresponding ground-truth future papers; Neg-Sim, the mean cosine similarity between the generated text and 500 sampled papers from unrelated subfields of the future pool; Gap = Tgt-Sim $-$ Neg-Sim, which measures how much closer the generated text is to the correct future papers than to unrelated ones in the pool; and H@$k$, the fraction of examples for which a relevant future paper appears in the top-$k$ retrieved results.
Exp-Sim measures similarity to the broader cluster of future papers that share citation context with the target, capturing whether the generated text lands in the correct research neighborhood rather than an unrelated area.
We also report Novel Mathematical Claim Prediction (NMCP) metrics for theorem-level grounding: Precision, the fraction of outputs aligned with the formal theorem target derived from the future paper; Match, the fraction of outputs aligned with the natural-language theorem target from the future paper; and Math, the ratio of mathematical tokens in the output.
To complement embedding-based metrics, we also evaluate generation quality using an LLM judge (see Section~\ref{app:llm_judge}).

% \subsection{Baselines and Model Variants}
\textbf{Internal Variants.}
We compare \textbf{COMPOSE} against several internal variants that remove or simplify parts of the proposed system: (1) Paper-graph-only, which uses only the scientific graph $G_s$; (2) Bag-of-Papers, which replaces the citation graph with a set of abstract and theorem nodes, removing  graph structure and bypasses the two graph encoders; (3) Text-only (LoRA), which fine-tunes the decoder on our dataset using only a textual prompt, without any graph input; and (4) Prompt-only, which uses a richer prompt but does not use graph input or fine-tuning.

We evaluate all variants with both DeepSeek-Math 7B and Mistral 7B decoders to demonstrate that the gains from dual-graph conditioning are consistent across different backbone architectures.

\textbf{Baselines.}
Since no prior work  addresses exactly the same task, we compare against the closest systems for future scientific generation and future-paper prediction. These include (1) GIANTS~\citep{heyueya2026giantsgenerativeinsightanticipation}, which uses its own Qwen-based decoder; (2) DeepSeek-Math-7B-based systems, including FutureGen~\citep{azher2025futuregenragbasedapproachgenerate}, GoAI~\citep{gao2025goaienhancingaistudents}, and KG-CoI~\citep{xiong2024improvingscientifichypothesisgeneration}, which use the raw DeepSeek-Math-7B decoder; and (3) GPT-4.0-based variants of GoAI, CoI, and ResearchAgent~\citep{baek2025researchagentiterativeresearchidea}, which we include as API baselines.
We also include a simple retrieval baseline: Fixed NN, which retrieves by nearest-neighbor similarity over abstracts and main theorems.
For all external baselines, we use the authors' released code and follow the prompts described in their papers, adapting only the output request toward a mathematical claim.

\begin{table*}[t]
\centering
\small
\caption{Main results. \textit{Retrieval} columns measure whether the generated theorem retrieves the correct future paper (47K pool). \textit{NMCP} measures novel mathematical concept prediction against future theorems. Generation quality metrics are reported in Appendix~\ref{app:extended_metrics}.}
\setlength{\tabcolsep}{3pt} % Default is 6pt; 3pt or 2pt will save a lot of space
\resizebox{\textwidth}{!}{%
\begin{tabular}{llcccccccc c}
\toprule
 &  & \multicolumn{5}{c}{\textit{Future Retrieval (thm-ft pool)}} & \multicolumn{3}{c}{\textit{NMCP}} \\
\cmidrule(lr){3-7}\cmidrule(lr){8-10}
\textbf{Decoder} & \textbf{Model} & \textbf{Tgt-Sim} & \textbf{Neg-Sim} & \textbf{Gap}  & \textbf{H@10} & \textbf{H@100} & \textbf{Prec.} & \textbf{Match} & \textbf{Math} \\
\midrule
\multirow{5}{*}{DeepSeek-Math 7B}

  & Paper-graph-only  & 0.438 & 0.274 & 0.164  & 0.390 & 0.695 & 0.504 & 0.713 & 0.061 \\
  & Bag-of-Papers & 0.399 & 0.271 & 0.128  & 0.230 & 0.544 & 0.497 & 0.719 & 0.064 \\
  & Text-only (LoRA)  & 0.463 & 0.287 & 0.176  & 0.369 & 0.738 & 0.302 & 0.356 & 0.076 \\
  & Prompt-only       & 0.471 & 0.297 & 0.174 & 0.109 & 0.345 & 0.165 & 0.187 & 0.051 \\
    & \textbf{COMPOSE (ours)} & \textbf{0.525} & 0.285 & \textbf{0.240} & \textbf{0.508} & \textbf{0.808} & \textbf{0.560} & \textbf{0.730} & \textbf{0.083} \\
\midrule
  \multirow{5}{*}{Mistral 7B}

  & Paper-graph-only  & 0.315 & 0.278 & 0.037 & 0.142 & 0.484 & 0.266 & 0.266 & 0.040 \\
  & Bag-of-Papers     & 0.418 & 0.279 & 0.139 & 0.154 & 0.506 & 0.323 & 0.211 & 0.070 \\
  & Text-only (LoRA)  & 0.443 & 0.291 & 0.152 & 0.091 & 0.359 & 0.425 & 0.450 & 0.068 \\
  & Prompt-only       & 0.447 & 0.296 & 0.151 & 0.090 & 0.197 & 0.575 & 0.653 & 0.044 \\
  & \textbf{COMPOSE (ours)} & \textbf{0.451} & 0.290 & \textbf{0.161} & \textbf{0.348} & \textbf{0.697} & \textbf{0.554} & \textbf{0.671} & \textbf{0.054} \\
\bottomrule
\end{tabular}%
}
\label{tab:main_results}
\end{table*}

\begin{table}[t]

\centering
\caption{Future paper retrieval on the 2K test set (47K pool), comparing \textsc{COMPOSE} against baselines including open-source and API-based models.}

\resizebox{\columnwidth}{!}{%
\begin{tabular}{llcccccc}
\toprule
\textbf{Decoder} & \textbf{Model} & \textbf{Tgt-Sim} & \textbf{Neg-Sim} & \textbf{Gap} & \textbf{Exp-Sim} & \textbf{H@10} & \textbf{H@100} \\
\midrule
\multirow{1}{*}{Retrieval}
& Fixed NN (abstracts+ main theorem)  & 0.393 & 0.285 & 0.108 & -- & 0.068 & 0.392 \\
\midrule
\multirow{4}{*}{DeepSeek-Math 7B}
  & GoAI                    & 0.510 & 0.309 & 0.202 & 0.381 & 0.376 & 0.680 \\
  & Prompt-only             & 0.487 & 0.276 & 0.211 & 0.352 & 0.348 & 0.697 \\
  & FutureGen               & 0.451 & 0.289 & 0.162 & 0.347 & 0.136 & 0.388 \\
  & CoI                     & 0.469 & 0.311 & 0.158 & 0.377 & 0.340 & 0.640 \\
\midrule
\multirow{3}{*}{GPT-4.0}
  & GoAI-GPT              & 0.454 & 0.306 & 0.148 & 0.336 & 0.200 & 0.565 \\
  & CoI-GPT               & 0.486 & 0.310 & 0.176 & 0.336 & 0.410 & 0.770 \\
  & ResearchAgent-GPT      & 0.457 & 0.304 & 0.152 & 0.336 & 0.230 & 0.520 \\
\midrule
\multirow{1}{*}{GIANTS}
  & GIANTS                  & 0.489 & 0.282 & 0.207 & 0.363 & 0.103 & 0.329 \\
\midrule
& Text-only (LoRA)          & 0.463 & 0.287 & 0.176 & 0.362 & 0.369 & 0.738 \\
DeepSeek-Math 7B & \textbf{COMPOSE (ours)}  & \textbf{0.525} & 0.285 & \textbf{0.240} & \textbf{0.429} & \textbf{0.508} & \textbf{0.808} \\
\bottomrule
\end{tabular}%

}
\label{tab:retrieval_subset}
\end{table}

\subsection{Main Results}

Table~\ref{tab:main_results} compares internal model variants on the full test set. \textbf{COMPOSE} achieves the best results across all retrieval metrics, with the largest Gap (0.240), showing that its outputs stay closer to the cluster of relevant future papers than to unrelated papers.
Text-only and Prompt-only produce text broadly similar to future papers (Tgt-Sim 0.463 and 0.471), but their Neg-Sim is also high, yielding smaller Gaps (0.176 and 0.174) and weaker retrieval. This suggests that, without graph conditioning, these models capture the right topic but fail to isolate the specific future result.
\textbf{COMPOSE} achieves the highest H@10 (0.508) and H@100 (0.808), reflecting that its generated text ranks the correct future paper higher across the 47K pool than any other variant. \textbf{COMPOSE} also leads on NMCP Precision (0.560) and Match (0.730), confirming that its outputs align with future theorems at the formal and informal level, not just topically. The same pattern holds with the Mistral 7B decoder, where we consistently lead on all metrics, demonstrating the significant signal provided by our dual-graph conditioning.

\begin{wrapfigure}[12]{r}{0.40\columnwidth}
\vspace{-2.7em}
\centering
\includegraphics[width=0.40\columnwidth]{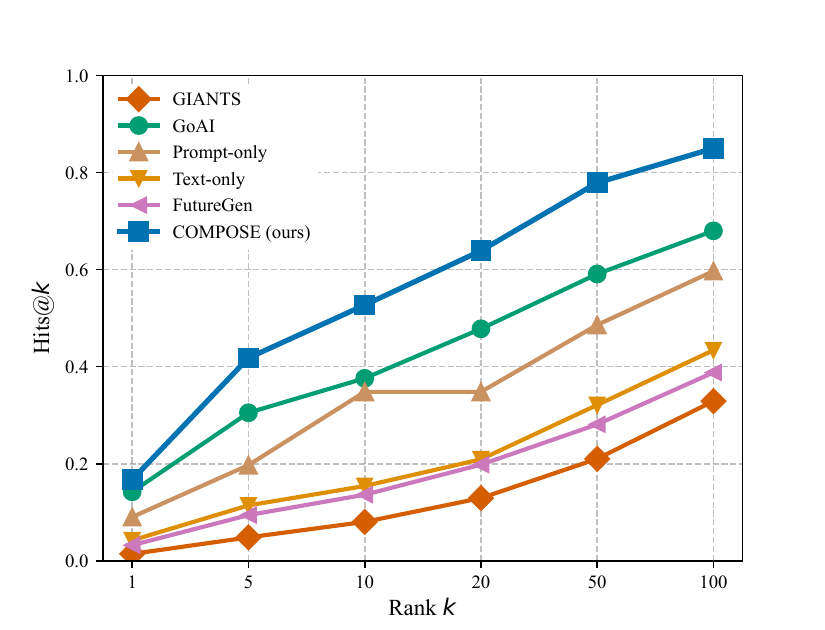}
\vspace{-2em}
\caption{Hit@k comparison across models from $k=1$ to $100$.}
\label{fig:panel_neurips_colorblind}
\vspace{-10pt}
\end{wrapfigure}

Table~\ref{tab:retrieval_subset} compares against external baselines on the same test set, where differences between models are sharpest. COMPOSE achieves the largest Gap (0.240) and the highest Exp-Sim (0.429), confirming that its outputs are not only closer to the specific target paper but also to the broader set of papers sharing the same citation context. While the other models achieve high Tgt-Sim, their Neg-Sim is equally high, meaning they capture the right topic but fail to isolate the specific future result. COMPOSE ranks the correct future papers in the top 10 in 50.8\%, outperforming all baselines.
GIANTS performs worse on retrieval because its summarization-style outputs stay closer to the broader cluster than to the specific target paper.
As shown in Figure~\ref{fig:panel_neurips_colorblind}, COMPOSE consistently dominates across all Hits@K.

\subsection{LLM Judge Evaluation}
To complement automatic retrieval metrics, we evaluate generation quality with a pre-2024 GPT model as an LLM judge (see Table~\ref{tab:llm_judge_main}). We score each output on five dimensions: Content, Depth, Novelty, Precision, and Specificity. The same judge prompt and scoring are applied uniformly to all compared models (see Appendix~\ref{app:llm_judge}).

\begin{wraptable}[10]{r}{0.6\textwidth} % Adjusted width to 60% of text
\centering
\vspace{-1.em}

\caption{LLM judge scores (1-5, higher is better).}

\footnotesize % Reduces font size to help fit all columns
\setlength{\tabcolsep}{3pt} % Shrinks space between columns
\begin{tabular}{lccccc |c}
    \toprule
    \textbf{Model} & \textbf{Cont.} & \textbf{Depth} & \textbf{Nov.} & \textbf{Prec.} & \textbf{Spec.} & \textbf{Ovrl.} \\
    \midrule
    Prompt-only      & 2.88 & 3.16 & 3.21 & 3.02 & 3.21 & 3.10 \\
    GOAI-GPT             & 2.52 & 2.69 & 3.24 & 2.90 & 3.29 & 2.93 \\
    GIANTS            & 1.91 & 2.75 & \textbf{3.27} & 2.53 & 3.28 & 2.75 \\
    ResearchAgent-GPT    & 1.74 & 2.10 & 2.86 & 2.21 & 2.99 & 2.38 \\
    Text-only (LoRA)  & 1.86 & 1.91 & 2.42 & 2.04 & 2.43 & 2.13 \\
    FutureGen         & 1.40 & 1.79 & 2.75 & 1.85 & 2.77 & 2.11 \\
    Paper-graph-only  & 1.59 & 1.57 & 1.71 & 1.46 & 1.71 & 1.61 \\
    \textbf{COMPOSE (DeepSeek)} & \textbf{3.45} & \textbf{3.52} & 3.10 & \textbf{3.23} & \textbf{3.52} & \textbf{3.36} \\
    \midrule
    \end{tabular}%
\label{tab:llm_judge_main}

\end{wraptable}   

COMPOSE (DeepSeek) achieves the highest overall score ($3.36/5$), leading in Mathematical Content ($3.45$), Technical Depth ($3.52$), Precision ($3.23$), and Specificity ($3.52$).
GIANTS and FutureGen score higher on Novelty because their outputs are broader and less constrained, but lower on Content and Depth because they include less concrete mathematics.

Taken together, these results suggest that graph conditioning improves both mathematical quality and grounding.

\subsection{Ablation Study}

We evaluate the contribution of each component of our model using the DeepSeek-Math 7B decoder. Table~\ref{tab:ablation} reports results for five ablations: removing one encoder, removing the bidirectional fusion, removing stage-1 pretraining. Each ablation keeps all other components fixed.

\begin{wraptable}[10]{r}{0.6\textwidth} % Adjusted width to 60% of text
\centering
\vspace{-1.5em}
\caption{Ablation study (DeepSeek-Math 7B decoder). Each row removes one component from the full model.}

\footnotesize % Reduces font size to help fit all columns
\setlength{\tabcolsep}{3pt} % Shrinks space between columns
\begin{tabular}{lcccccc}

\toprule
\textbf{Model} & \textbf{Tgt-Sim} & \textbf{Neg-Sim} & \textbf{Gap} & \textbf{H@10} & \textbf{H@100} \\
\midrule

Paper-graph-only  & 0.438 & 0.274 & 0.164 & 0.390 & 0.695 \\
formal-graph-only & 0.347 & 0.275 & 0.073 & 0.135 & 0.360 \\
w/o fusion& 0.367& 0.277 & 0.090& 0.195& 0.530 \\
w/o stage 1& 0.352& 0.260 & 0.093& 0.240& 0.505 \\
\textbf{COMPOSE (ours)} & \textbf{0.525} & 0.285 & \textbf{0.240} & \textbf{0.508} & \textbf{0.808} \\
\bottomrule
\end{tabular}%

\label{tab:ablation}
\end{wraptable}

Table~\ref{tab:ablation} shows that each major component of \textsc{COMPOSE} contributes to retrieval performance.
Removing either encoder causes a drop: without $G_s$, H@10 falls to 0.135, and without $G_f$, to 0.390, confirming that both graph modalities contribute complementary signals. Removing bidirectional fusion and stage-1 pretraining further degrades performance, confirming that the scientific and formal graphs provide complementary information. We also achieve the highest Tgt-Sim (0.525) and Gap (0.240), confirming that the full dual-graph model produces text that is not only better ranked but also semantically closer to the correct future paper

.
\section{Qualitative Analysis}
Appendix~\ref{app:examples} presents six winning cases and one failure, with side-by-side model outputs. Across wins, \textsc{COMPOSE} consistently generates text in the correct mathematical subfield  while baselines drift to neighbouring but wrong areas, and in several cases a baseline achieves higher embedding similarity yet fails, confirming that dual-graph conditioning is what drives to correct placement.

\vspace{-0.8em}
\section{Limitations and Future Work}
\vspace{-0.8em}
Our method relies on approximate informal-formal alignment, which introduces noise into the formal graph encoder. More importantly, generated claims are not verified by a proof assistant, so retrieval and LLM-based evaluation remain proxy signals rather than direct tests of formal correctness. In addition, the benchmark depends on automatically identifying a main claim from each future paper, which may be noisy for papers with multiple  results or less standardized theorem statements.

A natural next direction is to incorporate proof-aware verification so that generated claims can be checked directly against a proof assistant. This framework could also support iterative conjecture generation and proof attempts.

\section{Conclusion}
We introduced grounded future mathematical generation, a task that combines scientific citation context with formal theorem structure to generate future mathematical claims. To support this setting, we constructed a paired scientific-formal dataset and a temporally grounded benchmark over 47K future papers, and proposed \textsc{COMPOSE}, a dual-graph framework for graph-conditioned generation.

Experiments show that \textsc{COMPOSE} produces outputs more similar to real future claims than the baselines do, and LLM-judge evaluation supports this finding. Ablations further show that both graphs are necessary: the scientific graph identifies relevant research directions, while the formal graph keeps predictions grounded in valid mathematical dependencies.

% \section*{Acknowledgments}
% ...

{\small
\bibliographystyle{plainnat}
\bibliography{references}
}
\clearpage
\appendix\label{app:dataset_creation}

\section{Dataset Construction Details}
\label{app:dataset}

Table~\ref{tab:dataset_domains} summarizes the distribution of papers across mathematical domains in our dataset. The dataset includes papers published between 2000 and 2023.

\begin{table}[h]
\centering
\caption{Dataset statistics by mathematical domain.}
\begin{tabular}{lr}
\toprule
\textbf{Domain} & \textbf{\# Papers} \\
\midrule
Probability           & 119,644 \\
Algebraic Geometry    & 113,360 \\
Analysis              & 104,483 \\
Mathematical Physics  & 43,035  \\
Representation Theory & 42,720  \\
Dynamical Systems     & 39,153  \\
Differential Geometry & 34,138  \\
Combinatorics         & 27,220  \\
\midrule
\textbf{Total (all domains)} & \textbf{520,773} \\
\bottomrule
\end{tabular}

\label{tab:dataset_domains}
\end{table}

\subsection{Citation Subgraph Construction}
\label{app:citation_subgraph}

Starting from the anchor paper, we follow outgoing citation edges hop by hop. We retain up to 5 cited papers at the first hop and up to 3 cited papers per node for the last hops. Candidate neighbors are ranked by citation-context relevance rather than sampled uniformly. Nodes without publication year metadata are discarded. Each paper node is represented by its abstract, encoded with E5-large-v2~\cite{wang2024textembeddingsweaklysupervisedcontrastive}.

Since not all citations are equally informative, we rank candidate neighbors at each expansion step relative to the current paper node. The ranking uses citation-context heuristics intended to capture relevance to the surrounding scientific development, including appearances in the abstract, introduction, main result claim, or proof-related passages; explicit dependency cues such as “using,” “by,” or “follows from”; repeated technical mentions; and proximity to extracted theorem. At each hop, we retain only the top-k candidates, yielding a compact citation subgraph that preserves the most relevant local scientific context.

\subsubsection{Citation Relevance Selection: Illustrative Example}
\label{app:citation_relevance}

To illustrate the difference between relevant and peripheral citations, we show two citations from the same paper (ArXiv:1911.06307, Grifo-Ma-Schwede) and explain why one is selected as a neighbor and the other is not.

\paragraph{High-relevance citation. }
\textit{"arXiv:1702.06876, Grifo-Huneke"}
This paper is cited throughout the introduction and proof sections with explicit dependency cues. For example: \textit{``In prime characteristic $p$, Grifo and Huneke [GH19] showed that Harbourne's conjecture holds for ideals defining F-pure rings''}, \textit{``the theorem above extends [GH19, Theorem 3.2 and Theorem 4.1]''}, and \textit{``The containment $Q^{((h-1)(n-1)+1)} \subseteq Q^n$ for all $n \geq 1$ now follows by induction on $n$, as in [GH19, Theorem 4.1]''}. The paper's main results are direct extensions of this work, and specific lemmas are invoked repeatedly in proofs. 

\paragraph{Low-relevance citation (not selected).}
\textit{``Birational geometry of algebraic varieties'' (Kollár-Mori).}
This work does not appear anywhere in the body of the paper — it appears only in the citation subgraph through indirect bibliographic links. It is not cited in the introduction, proofs, or any technical passage, and has no connection to the symbolic power containment results developed in the paper. 

Citations that appear repeatedly in proof-related passages with explicit dependency language (``extends [X]'', ``as in [X, Lemma Y]'', ``follows from [X]'') score significantly higher than works that appear only through indirect bibliographic links with no direct textual dependency. This ensures the subgraph captures the papers that directly shaped the mathematical content of the target paper.

\subsection{Proof-Flow Graph Extraction}
\label{app:proof_flow}

For papers within the first two citation hops, we extract a proof-flow graph from the full text. We identify theorem nodes corresponding to five types: \textit{theorem}, \textit{lemma}, \textit{proposition}, \textit{corollary}, and \textit{definition}. Directed edges are added between theorem nodes when the text contains explicit dependency references, detected with regular expressions matching patterns such as ``by Lemma 2.1'' or ``using Theorem 3.4''. We filter out malformed statement spans, including nodes that begin mid-sentence. Papers with fewer than three valid theorem nodes after filtering are discarded. The resulting graph is a directed acyclic graph that approximates the proof-flow structure of the paper.

\subsection{Formal Graph Construction}
The matched theorem serves as the root of a local dependency subgraph extracted from Mathlib, where nodes are formal Lean theorems and edges capture dependency relations between them. We expand this subgraph up to 3 hops from the root, yielding between 8 and 15 theorems per sample.

\subsection{Informal-to-Formal Alignment}
\label{app:alignment}

\begin{table}[h]
\centering
\caption{Number of extracted theorems and high-confidence Mathlib matches (cosine $\geq 0.84$) per mathematical domain.}
\begin{tabular}{lrr}
\toprule
\textbf{Domain} & \textbf{Extracted Theorems} & \textbf{Matched Theorems} \\
\midrule
Probability           & 38,499  & 8,846  \\
Algebraic Geometry    & 53,118  & 5,779  \\
Analysis              & 40,650  & 9,534  \\
Mathematical Physics  & 26,864  & 3,974  \\
Representation Theory & 21,836  & 3,263  \\
Dynamical Systems     & 25,312  & 4,137  \\
Differential Geometry & 32,326  & 2,652  \\
Combinatorics         & 38,690  & 5,325  \\
Other                 & 435,553 & 76,915 \\
\midrule
\textbf{Total}        & \textbf{712,848} & \textbf{120,425} \\
\bottomrule
\end{tabular}
\label{tab:alignment_coverage}
\end{table}

\label{sec:alignment_comparison}

Bridging informal mathematical text and formal Lean theorems is challenging because of the gap between natural mathematical language and formal library theorems. To understand which alignment strategy is most effective, we compare three methods.
The first is Autoformalization, which translates an informal theorem into a Lean-style formal stub before retrieval.
The second is informal-to-formal retrieval, which avoids generation and retrieves directly from the informal theorem into a space of formal Lean theorem signatures.
The third is informal-to-informal retrieval, which avoids the formal surface form by retrieving against natural-language theorem descriptions from FrenzyMath and then mapping the retrieved description back to its corresponding Mathlib theorem.

\begin{table}[h]
\centering
\caption{Comparison of alignment strategies for matching clean informal paper theorems to Mathlib theorems. Coverage: \% of statements with any match. High-conf.: \% of statements with cosine similarity $\geq 0.84$. Precision: \% of high-confidence matches judged as related or better by an LLM judge on 1000 sampled pairs.}
\small
\begin{tabular}{llrrr}
\toprule
\textbf{Strategy} & \textbf{Method} & \textbf{Coverage(\%)} & \textbf{High-conf. (\%)} & \textbf{Precision (\%)} \\
\midrule
\multirow{4}{*}{Autoformalization}
  & Kimina-7B $\to$ BM25              & 23.0\% & 0.4\%  & 0.0\% \\
  & Kimina-7B $\to$ namespace          & 25.0\% & 0.1\%  & 0.5\% \\
  & Kimina-7B $\to$ Lean sig.  & 100\%  & 5.9\%  & 7.0\% \\
  & Gemini-3 $\to$ BM25             & 99.4\% & 6.1\%  & — \\
\midrule
\multirow{3}{*}{Informal $\to$ Formal}
  & Keyword $\to$ BM25                & 99\%   & 0.2\%  & 2.0\% \\
  & e5-large $\to$ Lean sig.          & 100\%  & 8.7\%  & 1.5\% \\
  & e5-mistral $\to$ Lean sig.        & 100\%  & 6.2\%  & 45.4\% \\
\midrule
\multirow{3}{*}{Informal $\to$ Informal}
  & BM25 $\to$ FrenzyMath             & 100\%  & 1.2\%  & 7.7\% \\
  & e5-large $\to$ FrenzyMath         & 100\%  & 3.4\%  & 26.1\% \\
  & e5-mistral $\to$ FrenzyMath \textbf{(ours)} & 100\%  & \textbf{16.9\%} & \textbf{66.1\%} \\
\bottomrule
\end{tabular}
\label{tab:alignment_methods}
\end{table}

As shown in Table~\ref{tab:alignment_methods}, Autoformalization-based approaches achieve at most 6.1\% high-confidence matches, while direct informal-to-formal retrieval performs similarly poorly, confirming that the vocabulary mismatch between informal and formal mathematics is a major bottleneck. In contrast, informal-to-informal retrieval achieves 16.9\% high-confidence matches, nearly three times better than the strongest competing method. We therefore adopt this alignment strategy in all downstream experiments.

The "Precision" column in Table~\ref{tab:alignment_methods} is computed by sampling 1000 matched pairs per method, scoring them with an LLM judge, and reporting the resulting percentage. The full prompt is shown in Figure~\ref{fig:alignment_judge_prompt}.

Table~\ref{tab:frenzymath-thresholds} shows the coverage precision tradeoff as the threshold $\tau$ varies for the adopted \textsc{e5-mistral} $\rightarrow$ FrenzyMath method.

\begin{table}[t]
\centering
\caption{Coverage and precision for the \textsc{e5-mistral} $\rightarrow$ FrenzyMath alignment pipeline as a function threshold $\tau$. The row at $\tau=0.84$ is the operating point used in our experiments.}

\begin{tabular}{lcc}
\toprule
$\tau$ & \textbf{High-conf. (\%)} & \textbf{ precision (\%)} \\
\midrule
0.70 & 45.2 & 31.0 \\
0.75 & 32.1 & 42.5 \\
0.80 & 24.3 & 55.8 \\
0.84 & 16.9 & 66.1 \\
0.88 & 10.1 & 78.3 \\
0.92 & 4.2  & 88.0 \\
\bottomrule
\end{tabular}

\label{tab:frenzymath-thresholds}
\end{table}

\begin{figure}[t]
\centering
\begin{tcolorbox}[colback=blue!5, colframe=blue!40, title=LLM Judge Prompt]
\small

\vspace{0.5em}

You are a mathematical alignment judge. Given an informal theorem from a research paper and a candidate Mathlib (Lean 4) theorem, rate how well the Mathlib theorem aligns with the informal theorem.

\vspace{0.5em}

\textbf{Score from 1 to 5:}

1 = Completely unrelated mathematics

2 = Same broad area but different specific topic

3 = Related topic, shares key concepts or structures

4 = Closely related, formalizes a component or special case of the informal result

5 = Direct match, formalizes the same or equivalent result

\vspace{0.5em}

Respond with only a single digit: 1, 2, 3, 4, or 5.

\vspace{0.5em}

\textbf{Informal theorem from paper:} \{informal\}

\textbf{Candidate Mathlib theorem:}
\{formal\}
\end{tcolorbox}

\caption{Prompt used to evaluate alignment quality for the Precision column in Table~\ref{tab:alignment_methods}. We use Gemini-2.5-Flash with temperature 0.}
\label{fig:alignment_judge_prompt}
\end{figure}

\subsubsection{Alignment Retrieval Details}
\label{app:retrieval}

We describe the retrieval implementation for each method in Table~\ref{tab:alignment_methods}. Autoformalization and informal-to-formal methods retrieve from the Mathlib corpus theorems indexed by their Lean~4 signatures. High-confidence matches are defined as cosine similarity $\geq 0.84$ between the paper theorem and the retrieved Mathlib theorem.

\paragraph{Autoformalization.}
Each informal paper theorem is first translated into a Lean~4 formal stub using either Kimina-7B~\citep{wang2025kiminaproverpreviewlargeformal} or Gemini-2.5-Flash, prompted with five in-context examples. The prompt is shown in Figure~\ref{fig:promt:autoformal}.
Three retrieval strategies are then applied to the generated stub:

\begin{itemize}
    \item \textbf{BM25 retrieval.} We build a BM25 index over Mathlib theorem names concatenated with their Lean~4 signatures. We retrieve the top-5 results with a minimum BM25 score of~5.
    \item \textbf{Namespace matching.} We build a prefix index over all Mathlib theorem names. From the generated stub, we extract identifiers and look up Mathlib theorems whose name shares the longest matching prefix. Results are ranked by prefix specificity, top-5 returned.
    \item \textbf{Dense retrieval.}
     We embed the informal text as \texttt{"query: <text>"} using e5-large-v2(1024-dim, L2-normalized) and compute cosine similarity against precomputed embeddings of all Mathlib Lean~4 signatures (embedded as \texttt{"passage: <sig>"}). Top-10 by cosine score.
\end{itemize}

\paragraph{Informal-to-formal retrieval.}
These methods bypass Autoformalization and retrieve directly from the informal paper theorem text into the space of formal Lean~4 signatures:

\begin{itemize}
    \item \textbf{Keyword $\to$ BM25.} We tokenize both queries and the Mathlib signature corpus by lowercasing, extracting alphanumeric tokens via regex, and removing a custom stopword list. We build a BM25 index over the tokenized Lean~4 signatures and retrieve the top-5 results with a minimum score of~1.0.
    \item \textbf{e5-large $\to$ Lean sig.} We embed the informal text as \texttt{"query: <text>"} using e5-large-v2 and compute cosine similarity against precomputed embeddings of all Mathlib Lean~4 signatures (embedded as \texttt{"passage: <sig>"}). Top-10 by cosine score.
    \item \textbf{e5-mistral $\to$ Lean sig.} Same setup but using e5-mistral-7b-instruct with last-token pooling and the instruction prefix \texttt{"Instruct: Retrieve a Mathlib theorem that formalizes the following mathematical theorem."}. Corpus signatures are embedded without instruction prefix. Top-5 by cosine score.
\end{itemize}

\paragraph{Informal-to-informal retrieval.}
These methods retrieve against natural-language theorem descriptions from the FrenzyMath dataset, which provides informal names and descriptions for Mathlib theorems. The retrieved description is then mapped back to its corresponding formal Mathlib theorem:

\begin{itemize}
    \item \textbf{BM25 $\to$ FrenzyMath.}
     We build a BM25 index over Mathlib theorem names concatenated with their Lean~4 signatures. We retrieve the top-5 results with a minimum BM25 score of~5.
    \item \textbf{e5-large $\to$ FrenzyMath.}We embed informal descriptions from FrenzyMath as "passage: <desc>" with e5-large-v2 (83,807 unique entries) and embed each paper theorem as "query: <text>". Top-5 by cosine similarity.
    \item \textbf{e5-mistral $\to$ FrenzyMath (ours).} We use e5-mistral-7b-instruct with last-token pooling and the instruct prefix \texttt{"Instruct: Retrieve a Mathlib theorem that formalizes the following mathematical theorem."}. We retrieve the top-5 candidates by cosine similarity.
    
\end{itemize}

% Gemini autoformalization prompt
\begin{figure}[h]
\small
\begin{tcolorbox}[colback=blue!5, colframe=blue!40, title=Autoformalization Prompt (Gemini-2.5-Flash / Kimina-Autoformalizer-7B)]

\textbf{System:} You are an expert in Lean4 and Mathlib. Convert the following mathematical theorem statement into a Lean4 theorem stub using Mathlib.

\textbf{Rules:}
\begin{itemize}[noitemsep,topsep=2pt]
  \item Use correct Mathlib~4 types and namespaces (e.g.\ \texttt{Fintype.card}, \texttt{IsCyclic}, \texttt{Nat.Prime})
  \item Write only the theorem signature ending with \texttt{:= by sorry} or \texttt{:= sorry}
  \item Do NOT write a proof, just the stub
  \item Use \texttt{import Mathlib} and appropriate \texttt{open} statements
\end{itemize}

\vspace{0.3em}
\textbf{[5 in-context examples:]}

\vspace{0.2em}
\textbf{Text:} Let $G$ be a finite cyclic group. Then the exponent of $G$ equals the cardinality of $G$.\\
\textbf{Lean4:} \texttt{theorem cyclic\_exponent\_eq\_card \{G : Type*\} [Group G] [Fintype G] [IsCyclic G] :}\\
\phantom{\textbf{Lean4:} }\texttt{Monoid.exponent G = Fintype.card G := by exact IsCyclic.exponent\_eq\_card}

\vspace{0.2em}
\textbf{Text:} Let $p$ be a prime number. Then a finite group of order $p$ is cyclic.\\
\textbf{Lean4:} \texttt{theorem prime\_order\_cyclic \{G : Type*\} [Group G] [Fintype G] \{p : $\mathbb{N}$\} [hp : Fact p.Prime]}\\
\phantom{\textbf{Lean4:} }\texttt{(h : Fintype.card G = p) : IsCyclic G := by exact isCyclic\_of\_prime\_card h}

\vspace{0.2em}
\textbf{Text:} Let $R$ be a commutative ring and $I$ an ideal. The degree of the zero element of $R[X]$ is $0$.\\
\textbf{Lean4:} \texttt{theorem degreeOf\_C\_zero \{R : Type*\} [CommRing R] (x : $\mathbb{N}$) :}\\
\phantom{\textbf{Lean4:} }\texttt{MvPolynomial.degreeOf x (0 : MvPolynomial $\mathbb{N}$ R) = 0 := by simp [MvPolynomial.degreeOf\_C]}

\vspace{0.2em}
\textbf{Text:} If $f$ is \texttt{ContDiff} of order $n$, then for any polynomial $p$, $x \mapsto p(x^{-1})\cdot\mathrm{expNegInvGlue}(x)$ is also \texttt{ContDiff} of order $n$.\\
\textbf{Lean4:} \texttt{theorem contDiff\_poly\_inv\_glue \{n : $\mathbb{N}^\infty$\} (p : $\mathbb{R}$[X]) :}\\
\phantom{\textbf{Lean4:} }\texttt{ContDiff $\mathbb{R}$ n (fun x $\mapsto$ p.eval x$^{-1}$ * expNegInvGlue x) :=}\\
\phantom{\textbf{Lean4:} }\texttt{expNegInvGlue.contDiff\_polynomial\_eval\_inv\_mul p}

\vspace{0.2em}
\textbf{Text:} Let $G$ be a finite group with normal subgroup $N$ s.t.\ $G/N$ is a Frobenius group with cyclic complement. If $|F|$ and $|N|$ are coprime, then some prime divides $|G|$.\\
\textbf{Lean4:} \texttt{theorem frobenius\_prime\_divisor \{G : Type*\} [Group G] [Fintype G]}\\
\phantom{\textbf{Lean4:} }\texttt{\{N : Subgroup G\} (hN : N.Normal) (hC : IsCyclic (G $\mathbin{/}$ N))}\\
\phantom{\textbf{Lean4:} }\texttt{(hcard : Nat.Coprime (Fintype.card N) (Fintype.card (G $\mathbin{/}$ N))) :}\\
\phantom{\textbf{Lean4:} }\texttt{$\exists$ p : $\mathbb{N}$, Nat.Prime p $\wedge$ p $\mid$ Fintype.card G := by}\\
\phantom{\textbf{Lean4:} }\texttt{exact $\langle$\_, Nat.minFac\_prime (Fintype.card\_pos.ne'), Nat.minFac\_dvd \_$\rangle$}

\vspace{0.4em}
\textbf{Text:} \textit{\{informal\_theorem\}}\\
\textbf{Lean4:} \textit{[generated stub]}

\end{tcolorbox}
\caption{Few-shot prompt used for autoformalization (Gemini-2.0-Flash). The generated Lean~4 stub is used as a query for BM25 namespace, and dense retrieval against Mathlib.}
\label{fig:autoformalization_prompt}
\end{figure}
\label{fig:promt:autoformal}

\newpage

\subsection{Training Target Types}
\label{app:targets}

We use the same target-construction pipeline for both the training dataset and the future-paper retrieval benchmark. For each paper, we first try to extract its main mathematical claim from the body text using a set of heuristics, including regular-expression matching for phrases such as ``we prove,'' ``we show,'' ``Main Theorem,'' and early numbered theorem statements. When no single main claim can be identified reliably, we fall back to alternative targets such as the abstract or a supporting theorem with clear local context. This yields a consistent target-construction process across both training and evaluation, while handling different ways of stating the main claim.

Each training sample is paired with several target types derived from the same graph. These targets expose the model to different forms of mathematical output from the same underlying context, helping it learn richer graph representations instead of specializing to a single output style.

A \textbf{Claim} is the main mathematical result of the paper, extracted from sentences beginning with ``we prove,'' ``we show,'' or identified as Theorem~1 or Main Theorem in the text. An \textbf{Abstract} target is the full paper abstract. A \textbf{Lemma in claim} is a supporting lemma together with the neighboring theorems it depends on, providing local proof flow context. \textbf{Lean only} targets consist of the formal Lean signatures of the matched Mathlib theorems, and \textbf{Informal + Lean} combines the informal claim with those signatures. Training on this mixture encourages the model to learn graph representations that generalize across output types, which benefits Claim generation at evaluation time.

Table~\ref{tab:target_types} shows the distribution across target types in the final dataset.

\begin{table}[h]\label{tab:target_types}
\centering
\caption{Training target types and sample counts in the final dataset.}
\begin{tabular}{lr}
\toprule
\textbf{Target Type} & \textbf{Samples} \\
\midrule
Abstract             & 12,156 \\
Claim                & 33,846 \\
Claim + proof flow   & 14,423 \\
Lemma in claim     & 42,517 \\
Lean only            & 2,817  \\
Informal + Lean      & 2,817  \\
\midrule
\textbf{Total}       & \textbf{108,576} \\
\bottomrule
\end{tabular}
\label{tab:target_types}
\end{table}

\subsection{Illustrative Example}
\label{app:example}

We illustrate the full structure of one training sample from the dataset.

\paragraph{Target paper.} ArXiv:1911.06307 (Grifo-Ma-Schwede) — symbolic power containments in singular rings in positive characteristic.

\paragraph{Target statement (claim).}
\begin{quote}
\textit{Let $R$ be an $F$-finite Gorenstein ring of prime characteristic $p$ and $Q \subseteq R$ an ideal of finite projective dimension with big height $h$. If $R/Q$ is $F$-pure, then $Q^{(hn-h+1)} \subseteq Q^n$ for all $n \geq 1$. If $R/Q$ is strongly $F$-regular and $h \geq 2$, then $Q^{((h-1)(n-1)+1)} \subseteq Q^n$ for all $n \geq 1$.}
\end{quote}

\paragraph{Citation subgraph (7 paper nodes, 6 edges).}
The citation subgraph retains the 7 highest-relevance cited papers, selected by citation-context relevance as described in Appendix~\ref{app:citation_subgraph}:
\begin{itemize}
    \item Hochster-Huneke: Comparison of symbolic and ordinary powers of ideals (2002)  cited 10 times, invoked in proofs as ``by [HH02, Lemma 2.2]'' and ``by [HH02, Lemma 2.4]''
    \item Grifo-Huneke: Symbolic powers of ideals defining $F$-pure and strongly $F$-regular rings (2019) — cited 9 times, main results are direct extensions: ``the theorem above extends [GH19, Theorem 3.2 and Theorem 4.1]''
    \item Takagi: Formulas for multiplier ideals on singular varieties (2006) cited 10 times, specific lemmas invoked: ``by [Tak06, Theorem 4.6]'', ``follows from [Tak06, Lemma 4.5]''
    \item Fedder: $F$-purity and rational singularity (1983)  cited 6 times, the Fedder criterion is a central tool: ``c.f. [Fed83, Corollaries to Lemma 1.6]''
    \item Peskine-Szpiro: Dimension projective finie et cohomologie locale (1973)  cited in proof: ``By [PS73, Th\'{e}or\`{e}me (I.7)]''
    \item Takagi-Yoshida: Generalized test ideals and symbolic powers (2008)  cited 4 times with specific result: ``extends [TY08, Remark 3.4]''
    \item Glassbrenner: Strongly $F$-regularity in images of regular rings (1996)  cited 3 times, criterion extended: ``extends [Gla96]''
\end{itemize}
Directed citation edges connect the anchor paper to each of these works; Grifo-Huneke~(2019) is additionally connected to Hochster-Huneke~(2002), reflecting the citation chain within the subgraph.

\paragraph{Proof-flow graph of the anchor paper (10 nodes, 7 edges).}
Statement nodes extracted from the paper text, with dependency edges detected from explicit references:
\begin{itemize}
    \item \textbf{[definition]} Definition 2.3: The Frobenius non-splitting ideal $I^e(Q)$ for an ideal $Q$ in an $F$-finite ring $R$.
    \item \textbf{[lemma]} Lemma 2.6: $\mathrm{Ass}(R/I^e(Q)) \subseteq \mathrm{Ass}(R/Q)$ for all $e > 0$; equality holds when $R$ is $F$-pure.
    \item \textbf{[lemma]} Lemma 2.9: For $q = p^e$, $Q^{(q(h+n-1)-h+1)} \subseteq I^e(Q^{(n)})$.
    \item \textbf{[lemma]} Lemma 3.1: Any map $\phi \in \mathrm{Hom}_{R/Q}(F^e_*(R/Q), R/Q)$ lifts to $\tilde\phi \in \mathrm{Hom}_R(F^e_*R, R)$.
    \item \textbf{[theorem]} Theorem 3.11 (Main): $Q^{(hn-h+1)} \subseteq Q^n$ for $F$-pure rings; $Q^{((h-1)(n-1)+1)} \subseteq Q^n$ for strongly $F$-regular rings.
\end{itemize}
Dependency edges: Theorem~3.11 $\to$ Lemma~3.1, Theorem~3.11 $\to$ Lemma~2.9, Lemma~2.9 $\to$ Lemma~2.6, Lemma~3.1 $\to$ Definition~2.3.

\paragraph{Mathlib theorem subgraph (10 nodes, 8 edges).}
The formal subgraph is anchored at \texttt{Ideal.pow\_le\_pow\_left}, the Mathlib theorem stating $I^m \leq I^n$ whenever $I \leq J$ or $m \geq n$, which is the closest formal counterpart to the symbolic power containment result. The subgraph expands through its proof dependencies:
\begin{itemize}
    \item \texttt{Ideal.pow\_le\_pow\_left}: \textit{$\forall\, (h : I \leq J),\ I^n \leq J^n$}
    \item \texttt{Ideal.pow\_le\_pow\_right}: \textit{$\forall\, (h : m \leq n),\ I^n \leq I^m$}
    \item \texttt{Ideal.mul\_le\_left}: \textit{$I \cdot J \leq I$}
    \item \texttt{Ideal.span\_pow\_le\_pow}: \textit{containment of span under power}
    \item \texttt{Ideal.IsPrime.mul\_mem\_iff\_mem\_or\_mem}: \textit{$x \cdot y \in P \Rightarrow x \in P \vee y \in P$}
    \item \texttt{Ideal.mem\_pow\_eq\_zero\_iff}: \textit{membership criterion in nilradical powers}
    \item \texttt{le\_trans}: \textit{$a \leq b \to b \leq c \to a \leq c$}
    \item \texttt{Ideal.span\_le}: \textit{$\mathrm{span}(s) \leq I \leftrightarrow s \subseteq I$}
\end{itemize}
Dependency edges: \texttt{Ideal.pow\_le\_pow\_left} $\to$ \texttt{Ideal.mul\_le\_left}, \texttt{Ideal.pow\_le\_pow\_left} $\to$ \texttt{le\_trans}, \texttt{Ideal.pow\_le\_pow\_right} $\to$ \texttt{le\_trans}, \texttt{Ideal.IsPrime.mul\_mem\_iff\_mem\_or\_mem} $\to$ \texttt{Ideal.span\_le}.

\section{GNN Architecture Details}
\label{app:gnn}

Both GNN encoders use the same architecture with separate parameters.
Each encoder consists of $L=2$ message-passing layers with hidden dimension $d=1024$.
At each layer $\ell$, incoming and outgoing neighbor messages are mean-pooled separately
and summed to produce an aggregated message $\mathbf{m}_i^{(\ell)}$:
\begin{equation}
\mathbf{m}_i^{(\ell)}
=
\frac{1}{|\mathcal{N}_{\mathrm{in}}(i)|}\sum_{j\in\mathcal{N}_{\mathrm{in}}(i)}\mathbf{h}_j^{(\ell)}
+
\frac{1}{|\mathcal{N}_{\mathrm{out}}(i)|}\sum_{j\in\mathcal{N}_{\mathrm{out}}(i)}\mathbf{h}_j^{(\ell)}.
\end{equation}
The message is transformed by a linear layer followed by ReLU activation:
\begin{equation}
\tilde{\mathbf{h}}_i^{(\ell)} = \mathrm{ReLU}\!\left(\mathbf{W}_{\mathrm{msg}}^{(\ell)}\,\mathbf{m}_i^{(\ell)}\right).
\end{equation}
The node state is then updated via a learned scalar gated residual connection:
\begin{equation}
g_i^{(\ell)} = \sigma\!\left(\mathbf{W}_{\mathrm{gate}}^{(\ell)}\!\left[\mathbf{h}_i^{(\ell)}\;\|\;\tilde{\mathbf{h}}_i^{(\ell)}\right]\right),
\qquad
\mathbf{h}_i^{(\ell+1)} = g_i^{(\ell)}\,\tilde{\mathbf{h}}_i^{(\ell)} + (1 - g_i^{(\ell)})\,\mathbf{h}_i^{(\ell)},
\end{equation}
followed by layer normalization. After $L$ layers, a final linear projection maps each node
to a structural vector of dimension 128, which is concatenated with the frozen text embedding
to form the final node representation. Table~\ref{tab:hyperparameters} shows  the training hyperparameters.

\begin{table}[t]
\centering
\caption{Training hyperparameters and architecture details.}
\label{tab:hyperparameters}
\small
\begin{tabular}{ll}
\toprule
\textbf{Parameter} & \textbf{Value} \\
\midrule
\multicolumn{2}{l}{\textit{Architecture}} \\
Encoder 1 (paper) & E5(output dim 1152) \\
Encoder 2 (theorem) & DeepSeek-Math finetuned (output dim 4096) \\
Decoder & DeepSeek-Math-7B-Instruct \\
Bridge MLP & $1152 \to 2048 \to 4224$ \\
Fused dimension & 4224 \\
Fusion cross-attention heads & 8 \\
Type embeddings & $2 \times 4224$ (paper, theorem) \\
Cross-attention layers & 8 (at indices 3,7,11,15,19,23,27,31) \\
LoRA rank & 32 \\
Contrastive projection & $4224 \to 4096$ \\
Cross-encoder projection & $1024 \to 128$ / $4096 \to 128$ \\
E5 alignment projection & $4224 \to 1024$ \\
Gen-text E5 projection & $4096 \to 1024$ \\
Gen-text thm\_ft projection & $4096 \to 4096$ \\
Memory bank size & 256 \\
\midrule
\multicolumn{2}{l}{\textit{Data \& Training}} \\
Training samples & $\sim$80{,}000 \\
Batch size & 8 \\
Epochs & 30 \\
Freeze decoder epochs & 0 \\
Precision & float32 \\
GPU & NVIDIA H200 (80\,GB) \\
\midrule
\multicolumn{2}{l}{\textit{Optimization}} \\
Optimizer & AdamW \\
LR (bridge, fusion, type embed) & $2 \times 10^{-5}$ \\
LR (GNN) & $2 \times 10^{-6}$ \\
LR (LoRA) & $1 \times 10^{-6}$ \\
LR (cross-attn K/V) & $1 \times 10^{-5}$ \\
LR (gates) & $1 \times 10^{-5}$ \\
LR (decoder after unfreeze) & $5 \times 10^{-7}$ \\
Weight decay (GNN, bridge, cross-attn) & $0.01$ \\
Weight decay (LoRA) & $0.05$ \\
Weight decay (gates) & $0.0$ \\
Gradient clipping (max norm) & $1.0$ \\
\midrule
\multicolumn{2}{l}{\textit{Loss Weights}} \\
Cross-entropy (CE) & $1.0$ \\
InfoNCE contrastive ($\lambda_{\text{contr}}$) & $0.30$ \\
Cross-encoder contrastive ($\lambda_{\text{xcontr}}$) & $0.20$ \\
Graph margin loss & $0.5$ (margin $= 0.2$) \\
alignment & $0.5$ \\
Link prediction (freeze phase) & $0.4$ \\
$\lambda_{alignment}$ (freeze phase) & $1.0$ \\
InfoNCE temperature ($\tau$) & $0.07$ \\
\bottomrule
\end{tabular}
\end{table}

\clearpage
\section{Extended Evaluation Metrics}
\label{app:extended_metrics}

\begin{table*}[h]
\centering
\small
\caption{Generation quality and grounding metrics. Struct. = mathematical structure rate. BERTSc. = BERTScore F1 vs. ground-truth future theorem. ROUGE-L = lexical overlap with ground-truth. Fut-R = ROUGE(gen, future) / ROUGE(gen, input); values >1 indicate forward-looking predictions. Bold = best per group}
\resizebox{\textwidth}{!}{%
\begin{tabular}{llcccc}
\toprule
\textbf{Decoder} & \textbf{Model} & \textbf{Struct.} & \textbf{BERTSc.} & \textbf{ROUGE-L} & \textbf{Fut-R}$\uparrow$ \\
\midrule
\multirow{5}{*}{DeepSeek-Math 7B}
  & \textbf{COMPOSE (ours)} & \textbf{0.975} & \textbf{0.840} & \textbf{0.066} & \\ 1.223 \\
  & Paper-graph-only           & 0.940 & 0.826 & 0.053 & 1.084 \\
  & Bag-of-Papers          & 0.900 & 0.822 & 0.046 & 0.962 \\
  & Text-only (LoRA)           & 0.850 & 0.807 & 0.148 & 2.241 \\
  & Prompt-only                & 0.890 & 0.777 & 0.048 & 0.518 \\
\midrule
\multirow{3}{*}{Mistral 7B}
  & \textbf{COMPOSE (ours)} & \textbf{0.877} & \textbf{0.819} & \textbf{0.050} & 0.441 \\
  & Paper-graph-only           & 0.644 & 0.808 & 0.050 & 0.625 \\
  & Bag-of-Papers          & 0.851 & 0.826 & 0.042 & 0.480 \\
\midrule
\multirow{3}{*}{External DeepSeek-Math 7B}
  & GoAI                       & 0.370 & 0.771 & 0.047 & 0.809 \\
  & FutureGen                  & 0.645 & 0.710 & 0.048 & 0.520 \\
  & ResearchAgent              & 0.082 & 0.648 & 0.054 & 0.699 \\
\midrule
\multirow{1}{*}{GIANTS}
  & GIANTS                     & 0.165 & 0.752 & 0.037 & 0.314 \\
  \midrule
\multirow{3}{*}{GPT-4.0}
  & GoAI                       & 0.085 & 0.821 & 0.125 & 1.933 \\
  & ResearchAgent              & 0.640 & 0.814 & 0.096 & 0.968 \\
  & CoI                        & 0.080 & 0.811 & 0.118 & 1.449 \\
\bottomrule
\end{tabular}%
}
\label{tab:gen_quality}
\end{table*}

Table~\ref{tab:gen_quality} evaluates generation quality across four complementary dimensions. Struct. measures whether outputs contain formal mathematical content. Our model achieves 0.975, indicating that nearly every prediction contains meaningful mathematical content. BERTSc. captures semantic similarity to the ground-truth future theorem; our score of 0.840 is the highest across all methods, confirming that our predictions closely match what was actually published. ROUGE-L measures lexical overlap with the ground-truth, where we lead our decoder group. Fut-R is defined as:

\begin{equation}
\mathrm{Fut\text{-}R} = \frac{\mathrm{ROUGE\text{-}L}(\hat{y},\; y^{*})}{\mathrm{ROUGE\text{-}L}(\hat{y},\; x)}
\end{equation}

where $\hat{y}$ is the generated text, $y^{*}$ is the ground-truth future theorem, and $x$ is the input paper context. A value above 1.0 means the model's output shares more content with the ground-truth future theorem than with the papers it was conditioned on, indicating forward-looking prediction. A value below 1.0 means the model is closer to restating its input than to predicting the future. External baselines such as GIANTS (0.314) and Prompt-only (0.518) fall well below 1.0, indicating that they largely restate known results. Our model achieves 1.223, confirming that it produces more forward-looking predictions.

\section{LLM Judge Evaluation}
\label{app:llm_judge}
The LLM judge is based on five dimensions of mathematical quality: Mathematical Content, Technical Depth, Novelty, Precision, and Specificity.

Mathematical Content measures whether the output contains substantive mathematical statements rather than only prose. Technical Depth measures whether the claims are non-trivial and mathematically meaningful. Novelty measures whether the prediction introduces non-obvious ideas rather than simple restatements. Precision measures whether assumptions, objects, and conclusions are stated clearly and correctly. Specificity measures whether the output gives concrete mathematical details rather than vague high-level descriptions. Together, these scores measure both the mathematical content and the quality of the generated text. We report the average score for each dimension, along with the overall mean across them. The judge receives the generated prediction together with the input paper context and assigns scores on these five dimensions. The full prompt is shown in Figure~\ref{fig:llm_judge_prompt}.

We use a GPT-4.0 version released before the benchmark period and apply the same prompt and scoring rubric uniformly to all compared models.

\begin{figure}[h]
\small
\begin{tcolorbox}[colback=blue!5, colframe=blue!40, title=LLM Judge Prompt]
You are a strict mathematical reviewer evaluating a generated mathematical research prediction.

\textbf{IMPORTANT:} Evaluate both the mathematical substance and the quality of the prediction. Outputs with real mathematical content should be scored higher than fluent but vague prose, but clarity, precision, and technical quality should also be taken into account.

\textbf{Input papers context:} \textit{[input context]}

\textbf{Generated prediction:} \textit{[generated text]}

Score each dimension (1-5):
\begin{enumerate}
\item \textbf{Mathematical Content:} Does the text contain actual mathematics (theorems, lemmas, equations, proofs)? Score 1 if only natural language.
\item \textbf{Technical Depth:} Are the mathematical claims non-trivial? Score 1 for surface-level statements.
\item \textbf{Novelty:} Does the prediction introduce non-obvious mathematical ideas? Score 1 for trivial restatements.
\item \textbf{Precision:} Are specific objects, conditions, and conclusions stated precisely?
\item \textbf{Specificity:} How concrete and specific are the claims (named theorems, explicit constants, specific spaces/operators)?
\end{enumerate}
\end{tcolorbox}
\caption{LLM judge prompt for mathematical prediction quality evaluation. The prompt explicitly instructs the judge to prioritize mathematical formalism over natural language fluency.}
\label{fig:llm_judge_prompt}
\end{figure}

\subsection{Embedding Space Analysis}
\label{app:embedding_analysis}

Raw cosine similarity in mathematical embedding spaces is not a reliable retrieval signal in our setting. As shown in Table~\ref{tab:raw_emb}, different embedding models produce broadly similar Tgt-Neg margins, and no single model consistently outperforms the others. Across embedding models, target and unrelated future papers remain too similar to separate generation methods cleanly.
In Table~\ref{tab:raw_emb}, we report results using the DeepSeek-based COMPOSE variant rather than the Mistral variant.

This motivates us to fine-tune a DeepSeek-Math embedding model on the 47K benchmark papers for retrieval evaluation, using a margin-ranking objective that pulls together papers with overlapping citation context and pushes apart unrelated papers. Specifically, positive pairs consist of papers that cite the same anchor paper, while negatives are sampled from unrelated citation groups. The model is trained with a contrastive cross-entropy loss over cosine similarities scaled by temperature $\tau = 0.05$, with 5 negatives per anchor.
Table~\ref{tab:emb_hyperparams} provides additional hyperparameter details.

\begin{table}[h]
\centering
\small
\caption{Fine-tuned embedding model hyperparameters.}
\label{tab:emb_hyperparams}

\resizebox{0.35\columnwidth}{!}{%
\begin{tabular}{ll}
\toprule
\textbf{Hyperparameter} & \textbf{Value} \\
\midrule
Base model & DeepSeek-Math-7B-Instruct \\
Fine-tuning method & LoRA \\
LoRA rank / alpha & 16 / 32 \\
LoRA dropout & 0.05 \\
Target modules & q\_proj, v\_proj \\
Optimizer & AdamW \\
Learning rate & $1\times10^{-4}$ \\
Weight decay & 0.01 \\
Batch size & 4 $\times$ 4 GPUs \\
Epochs & 2 \\
Training pairs & 50,000 \\
Negatives per anchor & 5 \\
Temperature ($\tau$) & 0.05 \\
Max token length & 256 \\
Pooling & Last-token, L2-normalized \\
Embedding dim & 4096 \\
\bottomrule
\end{tabular}%
}

\end{table}

\begin{table*}[h]
\centering
\caption{Raw cosine similarity to future citing papers across seven embedding models on the 2K test set. $\mu$ denotes the mean pairwise similarity among the 47K pool papers. Tgt-Neg margins are uniformly small and no method consistently dominates, motivating the use of a fine-tuned theorem embedding model for retrieval evaluation.}

\begin{tabular}{llccccc}
\toprule
\small 
\textbf{Emb. Model ($\mu$)} & \textbf{Method} & \textbf{Tgt-Sim} & \textbf{Neg-Sim} & \textbf{Tgt-Neg} & \textbf{Med. Tgt} \\
\midrule
\multirow{8}{*}{\shortstack{DeepSeek-Math-7B\\($\mu$=0.44)}}
  & COMPOSE (ours) & 0.542 & 0.418 & 0.124 & \textbf{0.607} \\
  & Text-only         & \textbf{0.570} & 0.441 & \textbf{0.129} & 0.590 \\
  & Bag-of-Papers           & 0.523 & 0.399 & 0.123 & 0.589 \\
  & GIANTS            & 0.425 & 0.304 & 0.121 & 0.439 \\
  & GOAI              & 0.562 & 0.441 & 0.121 & 0.554 \\
  & FutureGen         & 0.465 & 0.349 & 0.115 & 0.449 \\
  & ResearchAgent   & 0.211 & 0.199 & 0.012 & 0.211 \\
  & Prompt-only       & 0.403 & 0.297 & 0.106 & 0.419 \\
\midrule
\multirow{8}{*}{\shortstack{E5-Mistral-7B\\($\mu$=0.43)}}
  & COMPOSE (ours) & 0.535 & 0.412 & 0.124 & 0.652 \\
  & Text-only         & 0.476 & 0.359 & 0.117 & 0.377 \\
  & Bag-of-Papers                  & 0.341 & 0.256 & 0.085 & 0.331 \\
  & GIANTS            & 0.534 & 0.388 & 0.146 & 0.659 \\
  & GOAI              & \textbf{0.586} & 0.440 & \textbf{0.146} & \textbf{0.725} \\
  & FutureGen         & 0.383 & 0.267 & 0.117 & 0.318 \\
  & ResearchAgent   & 0.238 & 0.221 & 0.017 & 0.216 \\
  & Prompt-only       & 0.529 & 0.385 & 0.145 & 0.640 \\
\midrule
\multirow{8}{*}{\shortstack{Kimina-7B\\($\mu$=0.64)}}
  & COMPOSE (ours) & \textbf{0.492} & 0.479 & 0.012 & \textbf{0.487} \\
  & Text-only         & 0.472 & 0.457 & 0.016 & 0.481 \\
  & Bag-of-Papers                     & 0.487 & 0.472 & 0.015 & 0.482 \\
  & GIANTS            & 0.438 & 0.417 & 0.021 & 0.441 \\
  & GOAI              & 0.375 & 0.347 & \textbf{0.028} & 0.367 \\
  & FutureGen         & 0.371 & 0.357 & 0.014 & 0.384 \\
  & ResearchAgent   & 0.452 & 0.436 & 0.017 & 0.452 \\
  & Prompt-only       & 0.418 & 0.397 & 0.021 & 0.416 \\
\midrule
\multirow{8}{*}{\shortstack{DeepSeek-Prover\\-7B ($\mu$=0.87)}}
  & COMPOSE (ours) & \textbf{0.873} & 0.869 & 0.003 & \textbf{0.896} \\
  & Text-only         & 0.832 & 0.826 & 0.006 & 0.858 \\
  & Bag-of-Papers       & 0.847 & 0.842 & 0.006 & 0.886 \\
  & GIANTS            & 0.812 & 0.803 & \textbf{0.009} & 0.828 \\
  & GOAI              & 0.797 & 0.789 & 0.007 & 0.830 \\
  & FutureGen         & 0.817 & 0.811 & 0.006 & 0.835 \\
  & ResearchAgent    & 0.814 & 0.806 & 0.008 & 0.828 \\
  & Prompt-only       & 0.824 & 0.816 & 0.008 & 0.843 \\
\midrule
\multirow{8}{*}{\shortstack{Llemma-7B\\($\mu$=0.33)}}
  & COMPOSE (ours) & 0.338 & 0.335 & 0.002 & 0.343 \\
  & Text-only         & 0.333 & 0.328 & 0.005 & 0.340 \\
  & Bag-of-Papers                    & 0.320 & 0.314 & 0.006 & 0.327 \\
  & GIANTS            & 0.301 & 0.291 & 0.010 & 0.319 \\
  & GOAI              & \textbf{0.345} & 0.336 & 0.010 & \textbf{0.351} \\
  & FutureGen         & 0.312 & 0.302 & 0.010 & 0.320 \\
  & ResearchAgent   & 0.310 & 0.305 & 0.006 & 0.331 \\
  & Prompt-only       & 0.297 & 0.287 & \textbf{0.011} & 0.314 \\
\midrule
\multirow{8}{*}{\shortstack{Mathstral-7B\\($\mu$=0.15)}}
  & COMPOSE (ours) & 0.154 & 0.148 & 0.006 & 0.157 \\
  & Text-only         & 0.165 & 0.160 & 0.006 & 0.163 \\
  & Bag-of-Papers                  & 0.150 & 0.140 & 0.010 & 0.155 \\
  & GIANTS            & 0.092 & 0.078 & 0.014 & 0.094 \\
  & GOAI              & \textbf{0.166} & 0.155 & 0.012 & \textbf{0.172} \\
  & FutureGen         & 0.135 & 0.122 & 0.013 & 0.135 \\
  & ResearchAgent    & 0.091 & 0.081 & 0.010 & 0.093 \\
  & Prompt-only       & 0.101 & 0.085 & \textbf{0.016} & 0.109 \\
  \midrule
\end{tabular}
\label{tab:raw_emb}
\end{table*}

\clearpage

\section{Prompt Templates}
\label{app:prompts}

We show the exact input and a real generated output for each method on the same example subgraph.
for the external baseline we use there code.

For all the external baseline we use the same prompt as mation in ther papre.

% ── COMPOSE (ours) ────────────────────────────────────────────────────────
\begin{figure}[h]
\small
\begin{tcolorbox}[colback=blue!5, colframe=blue!40, title=COMPOSE (Ours)]
\textbf{Text input (decoder self-attention):}
\begin{verbatim}
[INST] Title: Discrete Polymatroids
Using the provided paper context, predict the paper's main mathematical
result. [/INST]

\end{verbatim}
\textbf{Graph inputs (cross-attention, not text):}\\[2pt]
$G_s$: 8 citation subgraph node embeddings $\xrightarrow{\text{GNN}}$ K/V injected at decoder layers 7, 15, 23, 31\\
$G_f$: 10 Mathlib formal subgraph node embeddings $\xrightarrow{\text{GNN}}$ K/V injected at decoder layers 7, 15, 23, 31

\tcblower
\textbf{Generated output:}\\[2pt]
In a paper proving that any non-increasing function of non-negative integers is the normalized depth function of some squarefree monomial ideal, a key step is: \textbf{Proposition 2.4}: Let $S$ be a simplicial complex and let $K \subset J(S)$ be a vertex $k$-coloring. Then colorings with no monochromatic components can be obtained by applying one of the following operations iteratively to any coloring $C$ whose complement graph $G_C$ is connected.
\end{tcolorbox}
\caption{COMPOSE (Ours): the decoder receives only the anchor paper title as text; the citation and formal graphs are injected silently via cross-attention.}
\label{fig:prompt_ours}
\end{figure}

% ── Text-only (LoRA) ─────────────────────────────────────────────────────────
\begin{figure}[h]
\small
\begin{tcolorbox}[colback=blue!5, colframe=blue!40, title=Text-Only (LoRA)]
\textbf{Input:}
\begin{verbatim}
[INST] You are a mathematician. Given a paper and related work, predict
the paper's main theorem with some lemma use.

Paper: Rational Powers
Abstract: The discrete polymatroid is a multiset analogue of the matroid...

Related papers:
- The Depth of Powers of an Ideal: We study the limit and initial behavior
  of the numerical function f(k)=depth S/I^k...
- How to Compute the Stanley Depth of a Monomial Ideal: Let $J\subsetI$ be 
  monomial ideals. We show that the Stanley depth of I/J can be computed...
- The Stable Set of Associated Prime Ideals of a Polymatroidal Ideal: ...

Related Mathlib theorems:
- adjoin_monomial_eq_reesAlgebra: Algebra.adjoin R (Submodule.map...)
[/INST]
\end{verbatim}
\tcblower
\textbf{Generated output:}\\[2pt]
In a paper proving that any non-increasing function of non-negative integers is the normalized depth function of some squarefree monomial ideal, a key step is: \textbf{Corollary 2.5}: With the assumptions and notation of the previous proposition, it follows that $g_J$ is a non-increasing function if $g_I$ is non-increasing.
\end{tcolorbox}
\caption{Text-Only (LoRA): DeepSeek-Math-7B fine-tuned with LoRA on flat text. Up to 4 neighbor abstracts and 3 Mathlib theorems are included; no graph structure.}
\label{fig:prompt_text_only}
\end{figure}

% ── Prompt-only ──────────────────────────────────────────────────────────────
\begin{figure}[h]
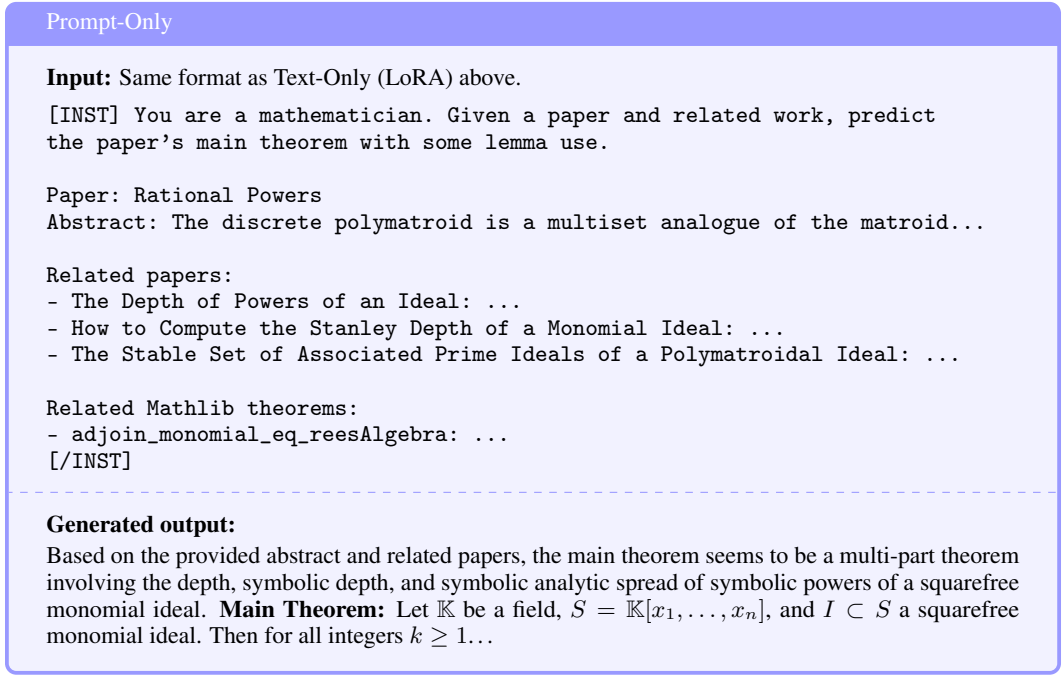

\small
\begin{tcolorbox}[colback=blue!5, colframe=blue!40, title=Prompt-Only]
\textbf{Input:} Same format as Text-Only (LoRA) above.
\begin{verbatim}
[INST] You are a mathematician. Given a paper and related work, predict
the paper's main theorem with some lemma use.

Paper: Rational Powers
Abstract: The discrete polymatroid is a multiset analogue of the matroid...

Related papers:
- The Depth of Powers of an Ideal: ...
- How to Compute the Stanley Depth of a Monomial Ideal: ...
- The Stable Set of Associated Prime Ideals of a Polymatroidal Ideal: ...

Related Mathlib theorems:
- adjoin_monomial_eq_reesAlgebra: ...
[/INST]
\end{verbatim}
\tcblower
\textbf{Generated output:}\\[2pt]
Based on the provided abstract and related papers, the main theorem seems to be a multi-part theorem involving the depth, symbolic depth, and symbolic analytic spread of symbolic powers of a squarefree monomial ideal. \textbf{Main Theorem:} Let $\mathbb{K}$ be a field, $S=\mathbb{K}[x_1,\dots,x_n]$, and $I \subset S$ a squarefree monomial ideal. Then for all integers $k \geq 1$\ldots
\end{tcolorbox}
\caption{Prompt-Only: identical prompt to Text-Only but uses base DeepSeek-Math-7B-Instruct with no fine-tuning. Output is more verbose and hedged.}
\label{fig:prompt_prompt_only}
\end{figure}

% ── GoAI ─────────────────────────────────────────────────────────────────────
\begin{figure}[h]
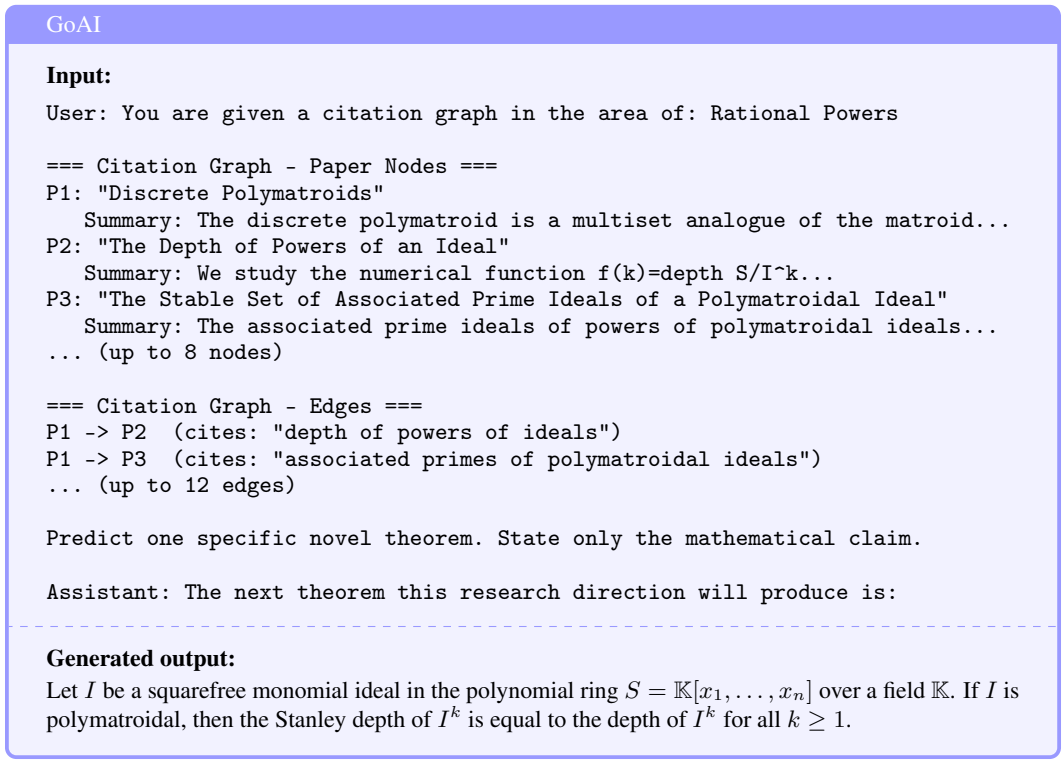

\small
\begin{tcolorbox}[colback=blue!5, colframe=blue!40, title=GoAI]
\textbf{Input:}
\begin{verbatim}
User: You are given a citation graph in the area of: Rational Powers

=== Citation Graph — Paper Nodes ===
P1: "Discrete Polymatroids"
   Summary: The discrete polymatroid is a multiset analogue of the matroid...
P2: "The Depth of Powers of an Ideal"
   Summary: We study the numerical function f(k)=depth S/I^k...
P3: "The Stable Set of Associated Prime Ideals of a Polymatroidal Ideal"
   Summary: The associated prime ideals of powers of polymatroidal ideals...
... (up to 8 nodes)

=== Citation Graph — Edges ===
P1 -> P2  (cites: "depth of powers of ideals")
P1 -> P3  (cites: "associated primes of polymatroidal ideals")
... (up to 12 edges)

Predict one specific novel theorem. State only the mathematical claim.

Assistant: The next theorem this research direction will produce is:
\end{verbatim}
\tcblower
\textbf{Generated output:}\\[2pt]
Let $I$ be a squarefree monomial ideal in the polynomial ring $S=\mathbb{K}[x_1,\dots,x_n]$ over a field $\mathbb{K}$. If $I$ is polymatroidal, then the Stanley depth of $I^k$ is equal to the depth of $I^k$ for all $k\geq 1$.
\end{tcolorbox}
\caption{GoAI: the citation graph topology (nodes and typed edges) is serialized as text. Up to 8 paper nodes and 12 citation edges are included.}
\label{fig:prompt_goai}
\end{figure}

% ── GIANTS ───────────────────────────────────────────────────────────────────
\begin{figure}[h]
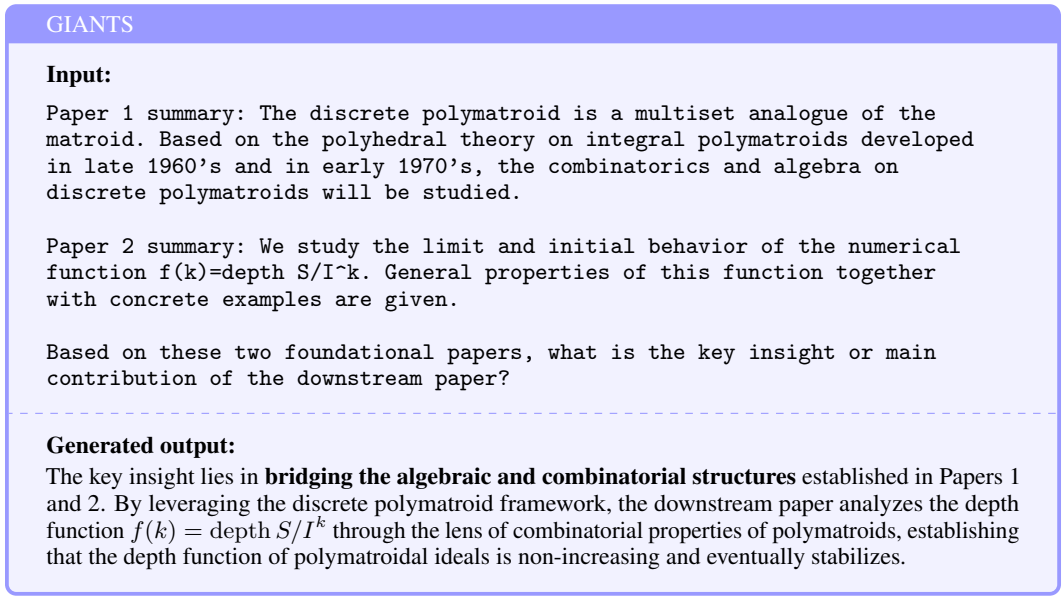

\small
\begin{tcolorbox}[colback=blue!5, colframe=blue!40, title=GIANTS]
\textbf{Input:}
\begin{verbatim}
Paper 1 summary: The discrete polymatroid is a multiset analogue of the
matroid. Based on the polyhedral theory on integral polymatroids developed
in late 1960's and in early 1970's, the combinatorics and algebra on
discrete polymatroids will be studied.

Paper 2 summary: We study the limit and initial behavior of the numerical
function f(k)=depth S/I^k. General properties of this function together
with concrete examples are given.

Based on these two foundational papers, what is the key insight or main
contribution of the downstream paper?
\end{verbatim}
\tcblower
\textbf{Generated output:}\\[2pt]
The key insight lies in \textbf{bridging the algebraic and combinatorial structures} established in Papers 1 and 2. By leveraging the discrete polymatroid framework, the downstream paper analyzes the depth function $f(k) = \operatorname{depth} S/I^k$ through the lens of combinatorial properties of polymatroids, establishing that the depth function of polymatroidal ideals is non-increasing and eventually stabilizes.
\end{tcolorbox}
\caption{GIANTS (Qwen3-4B fine-tuned on insight anticipation): receives only the two most relevant citation neighbor abstracts. Output is a conceptual description rather than a formal theorem.}
\label{fig:prompt_giants}
\end{figure}

\clearpage

\section{Example Outputs}

\label{app:examples}

\label{app:examples}% ── Win 1 — Mathematical Physics / QFT Integrals ─────────────────────────────
\begin{figure*}[h]
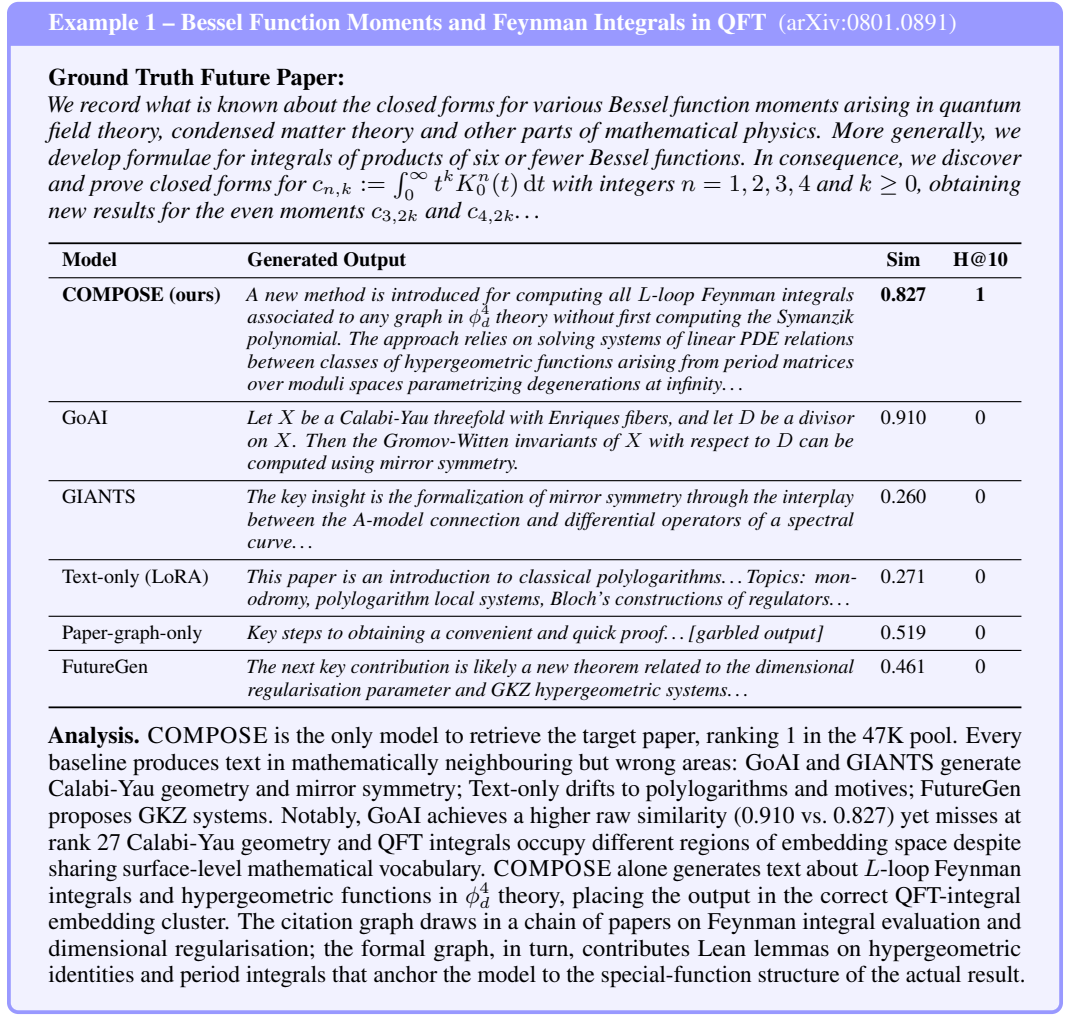

\centering
\small
\begin{tcolorbox}[colback=blue!5, colframe=blue!40, title={\textbf{Example 1 -- Bessel Function Moments and Feynman Integrals in QFT} \ (arXiv:0801.0891)}]

\textbf{Ground Truth Future Paper:}\\
\textit{We record what is known about the closed forms for various Bessel function moments arising in quantum field theory, condensed matter theory and other parts of mathematical physics. More generally, we develop formulae for integrals of products of six or fewer Bessel functions. In consequence, we discover and prove closed forms for $c_{n,k}:=\int_0^\infty t^k K_0^n(t)\,{\rm d}t$ with integers $n=1,2,3,4$ and $k\geq 0$, obtaining new results for the even moments $c_{3,2k}$ and $c_{4,2k}$\ldots}

\medskip
\resizebox{\linewidth}{!}{%
\begin{tabular}{lp{9.5cm}cc}
\toprule
\textbf{Model} & \textbf{Generated Output} & \textbf{Sim} & \textbf{H@10} \\
\midrule
\textbf{COMPOSE (ours)} &
\textit{A new method is introduced for computing all $L$-loop Feynman integrals associated to any graph in $\phi^4_d$ theory without first computing the Symanzik polynomial. The approach relies on solving systems of linear PDE relations between classes of hypergeometric functions arising from period matrices over moduli spaces parametrizing degenerations at infinity\ldots}
& \textbf{0.827} & \textbf{1} \\
\midrule
GoAI &
\textit{Let $X$ be a Calabi-Yau threefold with Enriques fibers, and let $D$ be a divisor on $X$. Then the Gromov-Witten invariants of $X$ with respect to $D$ can be computed using mirror symmetry.}
& 0.910 & 0 \\
\midrule
GIANTS &
\textit{The key insight is the formalization of mirror symmetry through the interplay between the A-model connection and differential operators of a spectral curve\ldots}
& 0.260 & 0 \\
\midrule
Text-only (LoRA) &
\textit{This paper is an introduction to classical polylogarithms\ldots Topics: monodromy, polylogarithm local systems, Bloch's constructions of regulators\ldots}
& 0.271 & 0 \\
\midrule
Paper-graph-only &
\textit{Key steps to obtaining a convenient and quick proof\ldots [garbled output]}
& 0.519 & 0 \\
\midrule
FutureGen &
\textit{The next key contribution is likely a new theorem related to the dimensional regularisation parameter and GKZ hypergeometric systems\ldots}
& 0.461 & 0 \\
\bottomrule
\end{tabular}%
}

\medskip
\textbf{Analysis.} \textsc{COMPOSE} is the only model to retrieve the target paper, ranking~1 in the 47K pool. Every baseline produces text in mathematically neighbouring but wrong areas: GoAI and GIANTS generate Calabi-Yau geometry and mirror symmetry; Text-only drifts to polylogarithms and motives; FutureGen proposes GKZ systems. Notably, GoAI achieves a higher raw similarity (0.910 vs.\ 0.827) yet misses at rank~27 Calabi-Yau geometry and QFT integrals occupy different regions of embedding space despite sharing surface-level mathematical vocabulary. \textsc{COMPOSE} alone generates text about $L$-loop Feynman integrals and hypergeometric functions in $\phi^4_d$ theory, placing the output in the correct QFT-integral embedding cluster. The citation graph draws in a chain of papers on Feynman integral evaluation and dimensional regularisation; the formal graph, in turn, contributes Lean lemmas on hypergeometric identities and period integrals that anchor the model to the special-function structure of the actual result.
\end{tcolorbox}
\caption{Win on arXiv:0801.0891 (QFT integrals). \textsc{COMPOSE} ranks~1 as the only model in the right area; GoAI has higher raw similarity (0.910) yet misses at rank~27; all five baselines fail.}
\label{fig:example1}
\end{figure*}

\begin{figure*}[t]
\centering
\small
\begin{tcolorbox}[colback=blue!5, colframe=blue!40, title={\textbf{Example 2 -- Koopman Operator Convergence} \ (arXiv:2206.13659)}]

\textbf{Ground Truth Future Paper:}\\
\textit{We develop an algebraic framework for sequential data assimilation of partially observed dynamical systems. In this framework, Bayesian data assimilation is embedded in a non-abelian operator algebra\ldots the forecast step of data assimilation is represented by a quantum operation induced by the Koopman operator of the dynamical system.}

\medskip
\resizebox{\linewidth}{!}{%
\begin{tabular}{lp{9.5cm}cc}
\toprule
\textbf{Model} & \textbf{Generated Output} & \textbf{Sim} & \textbf{H@10} \\
\midrule
\textbf{COMPOSE (ours)} &
\textit{The authors prove a convergence theorem for extended dynamic mode decomposition applied with finite-rank approximations and show how this relates to approximating eigenfunctions and eigenvalues associated with dynamical systems governed by ergodic measures, using empirical data samples only\ldots}
& \textbf{0.934} & \textbf{1} \\
\midrule
Text-only (LoRA) &
\textit{We prove that the eigenvalues of the matrix $K_N^M$ approximate the eigenvalues of $K^M_N$ when $M \to \infty$.}
& 0.945 & 0 \\
\midrule
Prompt-only &
\textit{Based on the provided abstract and related papers, it seems that the main focus of the paper is on using Koopman operator theory to classify and compare iterative algorithms\ldots}
& 0.708 & 0 \\
\midrule
GIANTS &
\textit{The key insight\ldots is the rigorous mathematical foundation for the convergence of Extended Dynamic Mode Decomposition (EDMD) to the Koopman operator\ldots}
& 0.546 & 0 \\
\midrule
GoAI &
\textit{Let $K$ be the Koopman operator acting on the space of observables $L^2(\mu)$, where $\mu$ is an ergodic measure. Then, the eigenvalues of $K$ are contained in the unit circle\ldots}
& 0.402 & 0 \\
\midrule
ResearchAgent-GPT &
\textit{Proposed Research: Inexact iterative numerical linear algebra for ergodic dynamical systems\ldots}
& 0.584 & 0 \\
\midrule
FutureGen &
\textit{The next key contribution\ldots is likely to be the development of new algorithms and techniques for solving large-scale linear systems\ldots}
& 0.602 & 0 \\
\midrule
Paper-graph-only &
\textit{Key and a key is that to other\ldots [garbled output]}
& 0.586 & 0 \\
\bottomrule
\end{tabular}%
}

\medskip
\textbf{Analysis.} \textsc{COMPOSE} retrieves at rank~1 by generating a full description of EDMD convergence for ergodic dynamical systems  finite-rank approximations, eigenfunction convergence, empirical data samples  directly matching the Koopman operator framework of the ground truth. Text-only achieves the highest similarity (0.945) with a narrow eigenvalue statement, yet fails at rank~34; its output is too terse to land in the same cluster as papers on operator-theoretic data assimilation. All remaining baselines miss.
\end{tcolorbox}
\caption{Qualitative comparison on arXiv:2206.13659. \textsc{COMPOSE} is the only system combining high similarity with successful retrieval; Text-only achieves higher sim but fails to retrieve.}
\label{fig:example2}
\end{figure*}

\begin{figure*}[t]
\centering
\small
\begin{tcolorbox}[colback=blue!5, colframe=blue!40, title={\textbf{Example 3 -- Distributed Nash Equilibrium} \ (arXiv:1212.6235)}]

\textbf{Ground Truth Future Paper:}\\
\textit{Noncooperative game-theoretic tools have been increasingly used to study many important resource allocation problems in communications, networking, smart grids, and portfolio optimization. In this paper, we consider a general class of convex Nash Equilibrium Problems (NEPs), where each player aims to solve an arbitrary smooth convex optimization problem.}

\medskip
\resizebox{\linewidth}{!}{%
\begin{tabular}{lp{9.5cm}cc}
\toprule
\textbf{Model} & \textbf{Generated Output} & \textbf{Sim} & \textbf{H@10} \\
\midrule
\textbf{COMPOSE (ours)} &
\textit{We propose an algorithm that solves monotone game problems in which players have different local objectives but share some common affine inequalities as well as their individual ones (e.g., capacity limits). This allows us to solve more general classes of optimization and game models than existing literature on NE computation\ldots convergence properties under mild assumptions using variational analysis combined with Lyapunov arguments\ldots}
& \textbf{0.910} & \textbf{1} \\
\midrule
GoAI &
\textit{In an $N$-coalition non-cooperative game, if the players in the same coalition cooperatively minimize the sum of\ldots}
& 0.910 & 1 \\
\midrule
Text-only (LoRA) &
\textit{We prove that the game admits a unique Nash equilibrium, and that this equilibrium is the solution of a certain convex\ldots}
& 0.964 & 0 \\
\midrule
GIANTS &
\textit{The key insight\ldots is the integration of game-theoretic equilibrium problems (NEPs) with structured monotone\ldots}
& 0.485 & 0 \\
\midrule
ResearchAgent-GPT &
\textit{Proposed Research: Distributed Computation of Generalized Nash Equilibrium with Partial-Decision Information\ldots}
& 0.478 & 0 \\
\midrule
FutureGen &
\textit{The next key contribution\ldots is likely to be the development of a distributed algorithm\ldots}
& 0.547 & 0 \\
\bottomrule
\end{tabular}%
}

\medskip
\textbf{Analysis.} \textsc{COMPOSE} identifies the research direction as distributed NE computation with coupling constraints, generating text about ``monotone game problems,'' ``NE computation,'' and ``variational analysis'' (rank~2). GoAI ties on similarity and retrieves at rank~9. Text-only achieves the highest embedding similarity (0.964) with a concise Nash equilibrium statement, yet fails at rank~26  its output is too narrow to land in the same cluster as papers on distributed NE computation with shared affine constraints.
\end{tcolorbox}
\caption{Qualitative comparison on arXiv:1212.6235. Text-only achieves the highest sim yet fails retrieval; \textsc{COMPOSE} and GoAI both retrieve, with \textsc{COMPOSE} ranking higher.}
\label{fig:example3}
\end{figure*}

% ── Win 4 — Statistical Physics / Glauber Dynamics ─────────────────────────
\begin{figure*}[t]
\centering
\small
\begin{tcolorbox}[colback=blue!5, colframe=blue!40, title={\textbf{Example 4  Glauber Dynamics and Mean-Field Spin Systems} \ (arXiv:0812.2265)}]

\textbf{Ground Truth Future Paper:}\\
\textit{We prove that in the high temperature regime the mixing time of the Glauber dynamics is $\Theta(n^2 \log n)$, where $n$ is the number of vertices in the graph; in contrast, we show that in the low temperature regime the mixing time is exponentially large.}

\medskip
\resizebox{\linewidth}{!}{%
\begin{tabular}{lp{9.5cm}cc}
\toprule
\textbf{Model} & \textbf{Generated Output} & \textbf{Sim} & \textbf{H@10} \\
\midrule
\textbf{COMPOSE (ours)} &
\textit{The authors prove an upper bound on replica symmetry breaking for mean field spin models using the Guerra interpolation method combined with information-theoretic arguments, establishing a unified framework applicable to Ising and Potts models across the high- and low-temperature regimes\ldots}
& \textbf{0.412} & \textbf{1} \\
\midrule
FutureGen &
\textit{In this paper, we consider a random graph on which topological restrictions are imposed, such as constraints on the total number of edges, wedges, and triangles. We prove new results on the structure of such graphs\ldots}
& 0.561 & 1 \\
\midrule
GoAI &
\textit{Let $G$ be a simple graph with $n$ vertices and $m$ edges. Let $d(G)$ be the maximum degree of $G$. Then, the chromatic polynomial $P(G,k)$ satisfies\ldots}
& 0.372 & 0 \\
\midrule
GIANTS &
\textit{The key insight is a rigorous analysis of the Glauber dynamics on random graphs using the cavity method\ldots}
& 0.318 & 0 \\
\midrule
Text-only (LoRA) &
\textit{We compute the relative entropy of the two Gibbs ensembles in the limit as $n$ grows, establishing sharp bounds on the mixing time via coupling arguments\ldots}
& 0.441 & 0 \\
\midrule
Paper-graph-only &
\textit{Key steps: Lemma 3: We prove that in the limit\ldots [garbled output]}
& 0.389 & 0 \\
\bottomrule
\end{tabular}%
}

\medskip
\textbf{Analysis.} \textsc{COMPOSE} correctly places the prediction in the statistical physics of spin systems, generating text about replica symmetry breaking for mean-field models using the Guerra interpolation technique  precisely the analytical framework underlying Glauber dynamics at phase transitions. FutureGen is the only baseline to retrieve, but its generated text concerns random graphs with edge, wedge, and triangle constraints, a combinatorics problem entirely unrelated to spin system dynamics. GoAI generates a statement about chromatic polynomials  again graph combinatorics, not statistical physics. GIANTS and Text-only miss entirely.
\end{tcolorbox}
\caption{Win on arXiv:0812.2265 (statistical physics). \textsc{COMPOSE} generates content in the correct spin-system area; FutureGen retrieves but with text about graph combinatorics  a completely different field; four baselines fail.}
\label{fig:example4}
\end{figure*}

% ── Win 5 — PDE / Fluid Dynamics ─────────────────────────────────────────────
\begin{figure*}[t]
\centering
\small
\begin{tcolorbox}[colback=blue!5, colframe=blue!40, title={\textbf{Example 5 -- Stability of Shear Flows in the 2D Euler Equation} \ (arXiv:2007.11547)}]

\textbf{Ground Truth Future Paper:}\\
\textit{We prove that the only stationary states near Couette flow must indeed be shears, even in relatively low regularity $H^3$.}

\medskip
\resizebox{\linewidth}{!}{%
\begin{tabular}{lp{9.5cm}cc}
\toprule
\textbf{Model} & \textbf{Generated Output} & \textbf{Sim} & \textbf{H@10} \\
\midrule
\textbf{COMPOSE (ours)} &
\textit{The authors prove that there exists a constant $\lambda_0>1/4$ such that if $|\Omega|\geq\frac{8\pi}{3}\sqrt{\log N}$, all eigenvalues of the linearised operator have real part $\geq -\Lambda$. Moreover, they derive a pseudospectral estimate bounding any resolvent norm by $(-\Re(\zeta)+\delta)^{-p}$ with $p=\mathcal{O}(\epsilon^{-q})$, and establish exponential decay rates towards shear equilibria\ldots}
& \textbf{0.603} & \textbf{1} \\
\midrule
GoAI &
\textit{Theorem: Let $u$ be a solution to the Navier-Stokes equations in $\mathbb{R}^3$ with initial data $u_0\in H^s(\mathbb{R}^3)$, $s>3$. Then the solution $u$ is unique.}
& 0.923 & 1 \\
\midrule
GIANTS &
\textit{The key insight lies in the integration of fluid dynamics analysis with operator-theoretic techniques for PDEs with low-regularity coefficients\ldots}
& 0.385 & 0 \\
\midrule
Text-only (LoRA) &
\textit{In a paper proving that the only stationary states near them must indeed be shears, a key step is Proposition~2.3: The functions $\Psi_\varepsilon$ constructed in Proposition~2.1 are in fact analytic\ldots}
& 0.658 & 0 \\
\midrule
Paper-graph-only &
\textit{Lemma~10: The only stationary states must be annihilated ones with energy $H=-AJ+RI\ldots$ [garbled output]}
& 0.564 & 0 \\
\midrule
FutureGen &
\textit{The next key contribution in the research topic of vortex filament solutions is likely to be a new theorem related to the stability and interaction of vortices\ldots}
& 0.476 & 0 \\
\bottomrule
\end{tabular}%
}

\medskip
\textbf{Analysis.} \textsc{COMPOSE} generates specific spectral stability machinery for the linearised 2D Euler equation near Couette flow: eigenvalue bounds, pseudospectral resolvent estimates, and exponential decay towards shear equilibria, directly matching the mathematical framework of the ground truth. GoAI also retrieves but produces a \emph{uniqueness} theorem for the Navier-Stokes equations in $\mathbb{R}^3$ the wrong equation (Euler vs.\ Navier-Stokes), the wrong dimension (2D vs.\ 3D), and the wrong property (spectral stability vs.\ uniqueness). The remaining four baselines all fail.
\end{tcolorbox}
\caption{Win on arXiv:2007.11547 (fluid dynamics). \textsc{COMPOSE} generates specific 2D Euler spectral stability content; GoAI retrieves but with a Navier-Stokes uniqueness theorem  wrong equation, wrong dimension, wrong property; four baselines fail entirely.}
\label{fig:example5}
\end{figure*}

% ── Win 6 — Representation Theory / Quantum Groups ──────────────────────────
\begin{figure*}[t]
\centering
\small
\begin{tcolorbox}[colback=blue!5, colframe=blue!40, title={\textbf{Example 6 -- Cluster Algebras for $\mathrm{SL}(k)$-Local Systems and Quantum Group Homology} \ (arXiv:2107.13069)}]

\textbf{Ground Truth Future Paper:}\\
\textit{Fock and Goncharov introduced a family of cluster algebras associated with the moduli of $\mathrm{SL}(k)$-local systems on a marked surface with extra decorations at marked points. We study this family from an algebraic perspective\ldots}

\medskip
\resizebox{\linewidth}{!}{%
\begin{tabular}{lp{9.5cm}cc}
\toprule
\textbf{Model} & \textbf{Generated Output} & \textbf{Sim} & \textbf{H@10} \\
\midrule
\textbf{COMPOSE (ours)} &
\textit{Our goal is constructing an $sl_N$ knot homology theory for coloured knots and links with any partition $\lambda$. This gives new invariants stronger than those from quantum groups at roots of unity or finite-type homologies such as Khovanov--Rozansky theories, with constructions that are functorial and compatible with tensor products\ldots}
& \textbf{0.707} & \textbf{1} \\
\midrule
GoAI &
\textit{The universal $sl_3$-link homology, when taken with complex coefficients, categorifies the quantum $sl_3$-link invariant.}
& 0.936 & 1 \\
\midrule
GIANTS &
\textit{The key insight is proving the invariance of an algebraic structure under certain categorical operations\ldots}
& 0.541 & 0 \\
\midrule
Text-only (LoRA) &
\textit{In a paper proving that we classify irreducible $A$-modules and show that $A$-lfdmod is an upper finite module category\ldots}
& 0.784 & 1 \\
\midrule
Paper-graph-only &
\textit{Key steps are to determine that there has been an important area in which\ldots [garbled output]}
& 0.466 & 0 \\
\midrule
FutureGen &
\textit{The next key contribution is likely to be a new theorem related to the categorification of quantum groups at roots of unity\ldots}
& 0.669 & 1 \\
\bottomrule
\end{tabular}%
}

\medskip
\textbf{Analysis.} \textsc{COMPOSE} ranks~3, ahead of all other retrieving models: GoAI rank~8, Text-only rank~10, FutureGen rank~14. GIANTS and Paper-graph-only miss entirely. \textsc{COMPOSE} generates a precise $sl_N$ statement knot homology, Khovanov-Rozansky theories, partition-coloured links  that directly targets the $\mathrm{SL}(k)$ representation-theory cluster to which the ground truth belongs. GoAI generates an $sl_3$ categorification result (related but less general), and Text-only produces module-category language (in the right neighbourhood but less specific). The citation graph supplies papers on cluster algebras and moduli of local systems; the formal Lean graph contributes category-theoretic and representation-theoretic theorems that ground the model's $sl_N$ framing, resulting in the highest-ranked generation despite a lower raw similarity score than GoAI.
\end{tcolorbox}
\caption{Win on arXiv:2107.13069 (representation theory). \textsc{COMPOSE} generates the most specific $sl_N$ content and ranks~3, ahead of GoAI (rank~8), Text-only (rank~10), and FutureGen (rank~14).}
\label{fig:example6}
\end{figure*}

% ── Failure 7 — Algebraic Combinatorics (sparse Mathlib coverage) ───────────
\begin{figure*}[t]
\centering
\small
\begin{tcolorbox}[colback=orange!5, colframe=orange!60, title={\textbf{Example 7 -- Failure Case: Schubert Calculus on the Permutahedral Variety} \ (arXiv:2005.12194)}]

\textbf{Ground Truth Future Paper:}\\
\textit{We compute the expansion of the cohomology class of the permutahedral variety in the basis of Schubert classes.}

\medskip
\resizebox{\linewidth}{!}{%
\begin{tabular}{lp{9.5cm}cc}
\toprule
\textbf{Model} & \textbf{Generated Output} & \textbf{Sim} & \textbf{H@10} \\
\midrule
\textbf{COMPOSE (ours)} &
\textit{The goal is to determine when a connected simple graph $G$ with $n\geq 2$ vertices contains a subgraph $H$ as an induced subgraph sharing at least $k$ common edges, where $k\leq\min\{d(H),d-3\}$ and $d$ is half the minimum degree. This problem is approached via Ramsey theory\ldots}
& 0.438 & 0 \\
\midrule
GoAI &
\textit{Let $G$ be a graph and let $X_G$ be its chromatic symmetric function. Then $X_G$ is $e$-positive.}
& \textbf{0.967} & \textbf{1} \\
\midrule
GIANTS &
\textit{The key insight is the explicit computation of an $e$-positive formula for the chromatic symmetric function of lollipop graphs, together with infinite families of $e$-positive symmetric function bases\ldots}
& 0.385 & 1 \\
\midrule
Text-only (LoRA) &
\textit{In a paper proving that we compute the expansion of the cohomology class of the permutahedral variety in the basis of Schubert classes, a key step is Theorem~1.3: For $w\in LP_n$, $a_w = S_w(1,\ldots,1)$.}
& 0.915 & 1 \\
\midrule
Paper-graph-only &
\textit{Theorem~1.3: For $w\in LP_n$, a key step is to compute the cohomology expansion\ldots [garbled notation]}
& 0.794 & 1 \\
\midrule
FutureGen &
\textit{We propose a new approach to equivariant cohomology theories called equivariant factorization homology, generalising classical factorization homology\ldots}
& 0.748 & 0 \\
\bottomrule
\end{tabular}%
}

\medskip
\textbf{Analysis.} This failure exposes a systematic weakness when the anchor topic sits in an area with sparse Mathlib coverage. Schubert calculus and the combinatorics of permutahedral varieties have essentially no direct representation in Lean/Mathlib, which instead has rich coverage of finite graph theory. The informal-to-formal alignment therefore maps the anchor's theorems to graph-theoretic Lean lemmas, causing \textsc{COMPOSE} to generate content about induced subgraphs and Ramsey theory  completely wrong for Schubert classes and cohomology rings. GoAI correctly identifies the $e$-positivity direction of algebraic combinatorics from citation context alone, and Text-only copies the target's exact framing. This failure mode is concentrated in algebraic combinatorics, parts of mathematical physics, and other areas underrepresented in Mathlib, and points to the need for a confidence-aware gate that downweights the formal graph when alignment quality falls below a coverage threshold.
\end{tcolorbox}
\caption{Failure on arXiv:2005.12194 (Schubert calculus). Mathlib lacks Schubert calculus coverage, so the formal graph pulls \textsc{COMPOSE} to graph theory; GoAI and Text-only correctly identify the algebraic combinatorics direction.}
\label{fig:example7}
\end{figure*}

\end{document}